%% file: main.tex
\documentclass{article}

\usepackage[final]{corl_2022} 

\usepackage[numbers]{natbib}
\usepackage{multicol}
\usepackage{multirow}
\usepackage{amsmath}
\usepackage{amssymb}
\usepackage{xcolor}
\usepackage{optidef}
\usepackage{bm}
\usepackage{todonotes}
\usepackage{wrapfig}
\usepackage{booktabs}
\usepackage{subcaption}
\include{mik_tools}
\usepackage{hyperref}
\usepackage[title]{appendix}
\usepackage{chngcntr}
\usepackage{float}

\title{Manipulation via Membranes: High-Resolution and Highly Deformable Tactile Sensing and Control}

\author{
Miquel Oller\hspace{15pt} Mireia Planas \hspace{15pt} Dmitry Berenson \hspace{15pt} Nima Fazeli\\
Department of Robotics, University of Michigan \\
Ann Arbor, MI 48109, United States\\
\texttt{\{oller, mireiap, dmitryb, nfz\}@umich.edu} \\
\url{https://www.mmintlab.com/manipulation-via-membranes}
}

\begin{document}
\maketitle


\vspace{-20pt}

\begin{abstract}
    Collocated tactile sensing is a fundamental enabling technology for dexterous manipulation. However, deformable sensors  
    introduce complex dynamics between the robot, grasped object, and environment that must be considered for fine manipulation. Here, we propose a method to learn soft tactile sensor membrane dynamics that accounts for sensor deformations caused by the physical interaction between the grasped object and environment. 
    Our method combines the perceived 3D geometry of the membrane with proprioceptive reaction wrenches to predict future deformations conditioned on robot action. Grasped object poses are recovered from membrane geometry and reaction wrenches, decoupling interaction dynamics from the tactile observation model.
    We benchmark our approach on two real-world contact-rich tasks: drawing with a grasped marker and in-hand pivoting. Our results suggest that explicitly modeling membrane dynamics achieves better task performance and generalization to unseen objects than baselines.
\end{abstract}

\keywords{Manipulation, tactile control, deformable tactile sensors} 



\input{sections/Introduction}

\input{sections/RelatedWork}

\input{sections/ProblemStatement}

\input{sections/Methodology}

\input{sections/Experiments}

\input{sections/Discussion}


\clearpage
\acknowledgments{
This research project is supported by Toyota Research Institute under the University Research Program (URP) 2.0. 
This work has been partially supported by the mobility grants program of Centre de Formació Interdisciplinària Superior (CFIS) - Universitat Politècnica de Catalunya (UPC). We would also like to show our gratitude to the anonymous reviewers for their helpful comments in reviewing the paper. 
We also thank the members of the Manipulation and Machine Intelligence (MMINT) Lab for their support and feedback.

}


\small\bibliography{references}  

\input{sections/Supplementary} 

\end{document}

%% file: mik_tools.tex
\newcommand{\R}[0]{\mathbb{R}} 

\renewcommand{\vec}[1]{\bm{#1}}
\newcommand{\predvec}[1]{\hat{\vec{#1}}}

\newcommand{\mat}[1]{\bold{#1}}

\newcommand{\taskstate}[0]{\vec{x}}
\newcommand{\state}[0]{\vec{s}}
\newcommand{\dynstate}[0]{\state}

%% file: sections/Introduction.tex
\vspace{-5pt}
\section{Introduction}
\vspace{-5pt}
\begin{wrapfigure}[18]{r}{0.6\textwidth}
    \vspace*{-10pt}
     \centering
    \includegraphics[width=0.6\textwidth]{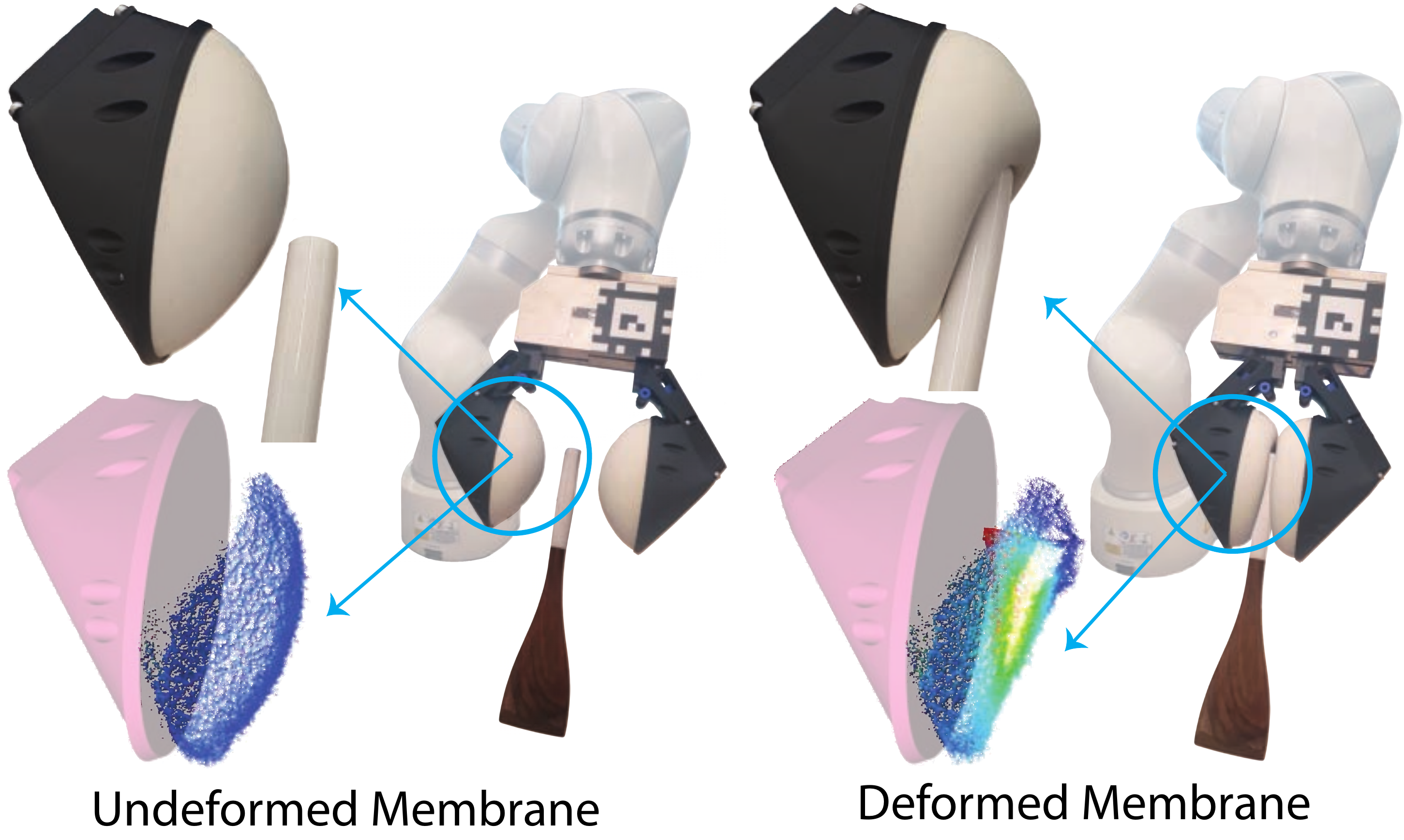}
    \caption{\textbf{Membrane Deformation Visualization} The sensor membranes deform significantly as a result of their interaction with the grasped object and environment. (top) real deformation, (bottom) perceived deformation.}
    \label{fig:teaser}
\end{wrapfigure}

Tactile sensing collocated at the contact interface between the robot and environment is a key enabling technology for dexterous manipulation \cite{rodriguez2021unstable}. The resulting information-rich and highly discriminative tactile cues are far more informative than proprioceptive external joint torques. To date, a number of high-resolution collocated tactile sensors have gained traction in the robotics community including Soft Bubbles \cite{alspach2019soft}, GelSlim \cite{donlon2018gelslim}, DIGIT \cite{lambeta2020digit}, and FingerVision \cite{yamaguchi2018fingervision}. However, there are two coupled challenges in using this class of sensors: the high-dimensionality of the sensor signal (signature) and the sensor dynamics introduced between the robot and what it is contacting. 

Recent progress in exploiting these high-dimensional signatures has resulted in sensor specific algorithms for state-estimation \cite{bauza2022tac2pose, bauzaFirstTouch, daolinGelslimContactSensing, bubblesStateEstimation, kelestemur2022tactile, narang2021interpreting} and controls \cite{li2013control, lepora2017exploratory, tactileMPC, van2016stable}. However, most progress has addressed low-deformation sensors with negligible dynamics (e.g., GelSlim, DIGIT, and FingerVision). These dynamics cannot be ignored for high-deformation tactile sensors (e.g., Soft Bubbles). This is because large sensor deformations lead to both non-negligible relative motion of the grasped object with respect to the end-effector and incipient slip as the object is brought into contact with the environment. When used effectively, this high compliance is desirable for contact rich interactions as it allows for large and stable contact patches as well as gradual force build up.

In this paper, we first illustrate deformation and force transmission differences between hard and soft tactile sensors by showing the relative behavior of the Soft Bubbles and GelSlim 3.0. The mechanical interface of the former is an inflated latex membrane while the latter uses a dense polymer. This exercise motivates our main contribution: a method to learn soft tactile sensor membrane dynamics that accounts for the sensor deformations caused by the physical interaction between the grasped object and environment. Our approach integrates the perceived 3D geometry of the sensor membrane with proprioceptive reaction wrenches and predicts future membrane deformations conditioned on robot actions. A key feature of our method is the decoupling of the sensor membrane dynamics from the tactile observation model. Tactile observation models compute features such as the object pose and extrinsic contact location from the sensor state (3D geometry and reaction wrenches). This decoupling exploits the fact that sensor membrane dynamics are shareable across tasks because they are inherent to the sensor mechanics while the observation model can be task-specific. We demonstrate how our method enables reasoning over the joint dynamics of the robot, grasped object, and environment to enable precise control of object pose and force transmission and empirically compare our method against 5 baselines on two real-world contact rich robotic tasks.

%% file: sections/RelatedWork.tex
\vspace{-10pt}
\section{Related Work}
\vspace{-10pt}

There are two major types of tactile sensing: localized and distributed. Localized sensing, here referring to the use of a single force-torque sensor often at the robot wrist, summarizes external contact information as a wrench composed by six numbers, 3 linear and 3 torque terms. Localized sensing can be provided by the robot through joint torques or using F/T sensors \cite{de2006collision,magrini2014estimation,manuelli2016localizing}. In contrast to localized, distributed tactile sensing collocated at the contact interface can provide information dense feedback in the form of images \cite{alspach2019soft,donlon2018gelslim,yamaguchi2018fingervision} or pressure distributions \cite{biotak}. The survey by \citet{yamaguchi2019recent} summarizes recent advances in these sensors. 
With the promise of information dense sensing; however, comes challenges including novel dynamics at the contact interface, and tactile signature representation for state-estimation, prediction, and controls.

Finding good tactile sensing representations is challenging because it typically requires either expert knowledge or considerable amounts of data. Approaches such as TACTO \cite{tacto} and \citet{narang2021interpreting} have reduced data requirements by simulating DIGIT \cite{lambeta2020digit} and SynTouch Biotac pressure-based sensors \cite{biotak}, respectively. Building on this, \citet{tarik2022} propose an object pose estimator trained in simulation that transfers to the real-world. 
However, current tactile simulators only support a small set of relatively rigid sensors and simulation of soft sensor, much like other deformables, remains a significant challenge. Therefore, our data is collected from real-world interactions.

Early work exploited dense tactile signals in closed-loop controllers for contact servoing \cite{li2013control} or active exploration \cite{lepora2017exploratory}. More recently, tactile control approaches have been used in robotic bi-manual object manipulation \cite{hogan2020tactile}, deformable object manipulation \cite{she2021cable}, and peg-in-hole insertion \cite{kim2021active}. These later approaches use the GelSlim tactile sensor \cite{donlon2018gelslim} which exhibits small and approximately linear deformations. This simplified model is not effective for the Soft Bubbles and assuming sticking  external contact is restrictive in many applications (see Sec.~\ref{sec:comparison} and \ref{sec:experiments}). 

Closely related to our work, previous work has explored explicitly modelling the tactile sensor dynamics. \citet{van2016stable} projected the Biotak tactile signals into a learned low-dimensional representation where dynamics are linear. Other approaches such \citet{tactileMPC} have modelled the dynamics of a vision-based tactile sensor such as GelSight. Their entire control pipeline is formulated in the sensor space, requiring the goal states to be specified as tactile states. Instead, we decouple the sensor dynamics from the task dynamics, allowing us to specify tasks in terms of object poses and forces transmitted to the environment. \citet{lambeta2020digit} extended their control formulation to a multi-finger control setting by adding structure to the tactile dynamics predictions. This approach conditions the dynamics on the estimated object pose, which creates a dependency on the pose estimation and may limit the model's ability to generalize to different objects.

%% file: sections/ProblemStatement.tex
\vspace{-5pt}
\section{Problem Formulation} \label{sec:prob-form}
\vspace{-5pt}


The goal of the robot is to control the grasped object pose and the force transmitted to the environment as the object is brought into contact. Our central idea is to represent these ``task dynamics’’ by modeling them using the tactile sensor membrane dynamics. We are motivated by the fact that the grasped object pose and transmitted forces are resolved simultaneously with the membrane deformation and end-effector reaction wrenches. This means the robot can plan for desired task states by reasoning through the membrane dynamics -- i.e., planning actions that result in membrane deformations that in turn result in the desired poses and transmitted forces.

\begin{wrapfigure}[15]{r}{0.25\textwidth}
    \vspace*{-2pt}
    \centering
    \includegraphics[width=0.25\textwidth]{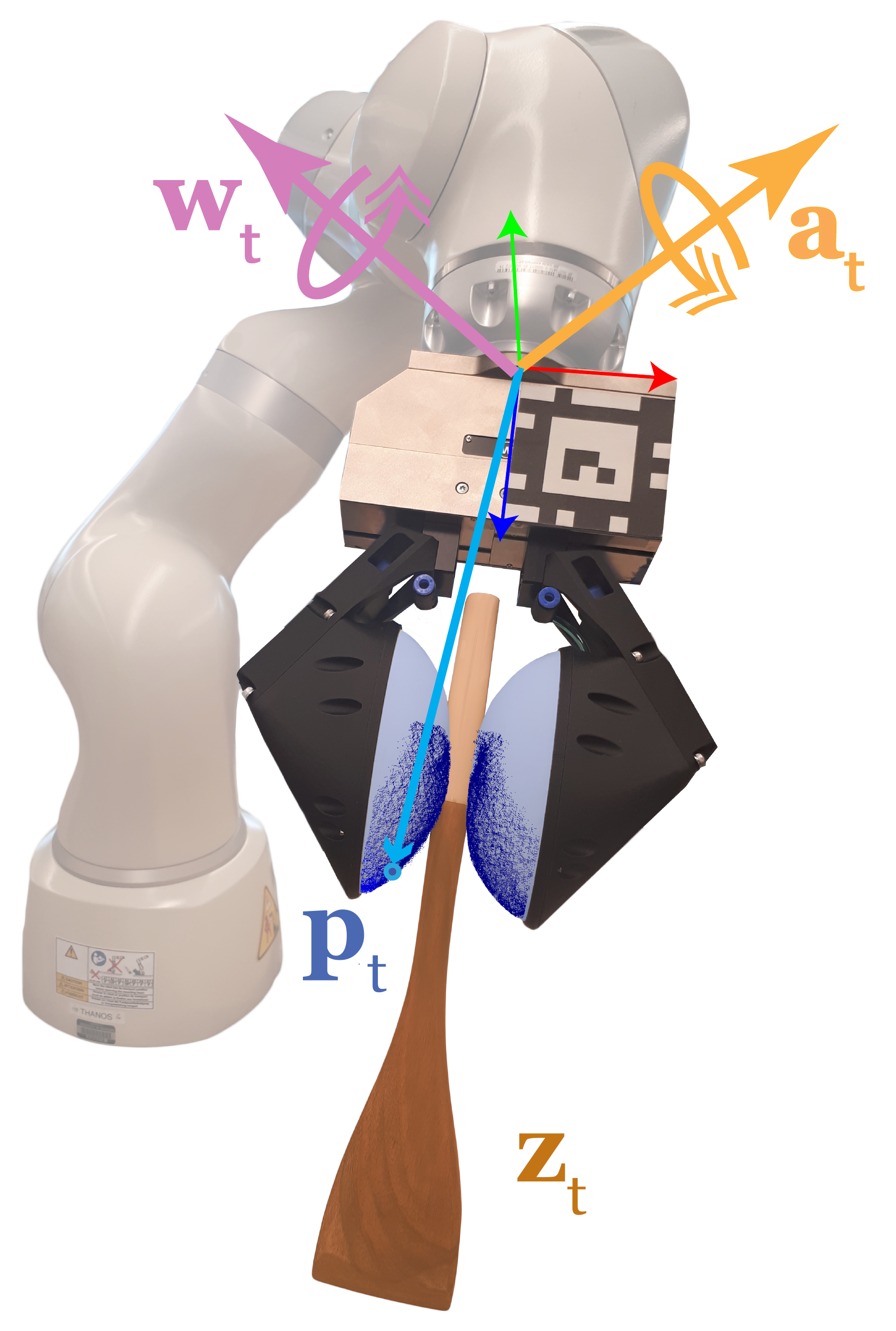}
    \caption{$\vec{p}_t$: membrane state, $\vec{w}_t$: external wrench, $\vec{z}_t$: object geometry, $\vec{a}_t$: robot action.}
    \label{fig:prob-form}
\end{wrapfigure}

Let $\vec{p}_t$ denote the 3D pose of any points on the surface of the membrane, then the dynamics model $\vec{p}_{t+1}=f(\vec{p}_t, \vec{w}_t, \vec{a}_t, \vec{z}_t)$ represents how each point deforms conditioned on the current membrane geometry, end-effector reaction force $\vec{w}_t$, robot action $\vec{a}_t$, and the grasped object $\vec{z}_t$. Next, let $\vec{q}_t$ denote the object-environment configuration, then the observation model $\vec{q}_t = g(\vec{p}_t, \vec{w}_t, \vec{z}_t)$ maps the current membrane geometry and reaction force to configurations. We define the task state as $\taskstate_t = (\vec{q}_t, \vec{w}_t)$ and goal as $\taskstate_g$. Planning and control reduces to solving the optimization problem:
\begin{mini*}|s| 
{\vec{a}_t}{ \sum_{t=0}^{N} (\taskstate_t-\taskstate_g)^{T} \mat{Q} (\taskstate_t-\taskstate_g)+ \vec{a}_t^{T} \mat{R} \vec{a}_t}
{}{}
\addConstraint{\vec{p}_{t+1}, \vec{w}_{t+1} = f(\vec{p}_t, \vec{w}_t, \vec{a}_t, \vec{z}_t), \quad \vec{q}_{t}=g(\vec{p}_t, \vec{w}_t, \vec{z}_t)}
\end{mini*} 
where $\mat{Q}$ and $\mat{R}$ are positive definite matrices weighing the relative importance of reaching goal states with effort. The key to solving this problem is learning a sufficiently-accurate estimate of $f$, which is the focus of this paper.

%% file: sections/Methodology.tex
\vspace{-5pt}
\section{Methods}
\vspace{-5pt}

In this section, we first illustrate characteristic deformation and force transmission differences between soft and hard sensors (Soft Bubbles vs. GelSlim 3.0) to motivate the need for our membrane dynamics. Next, we present our main contribution: our approach to learning a soft tactile sensor membrane dynamics model that accounts for the sensor deformations caused by the physical interaction between the grasped object and environment as well as an observation model used to extract the task state. Finally, we describe the controller which leverages these dynamics for task execution.

\vspace{-5pt}
\subsection{Sensor Deformation Evaluation} \label{sec:comparison}
\vspace{-5pt}

\begin{figure}[hb]
    \centering
    \includegraphics[width=\textwidth]{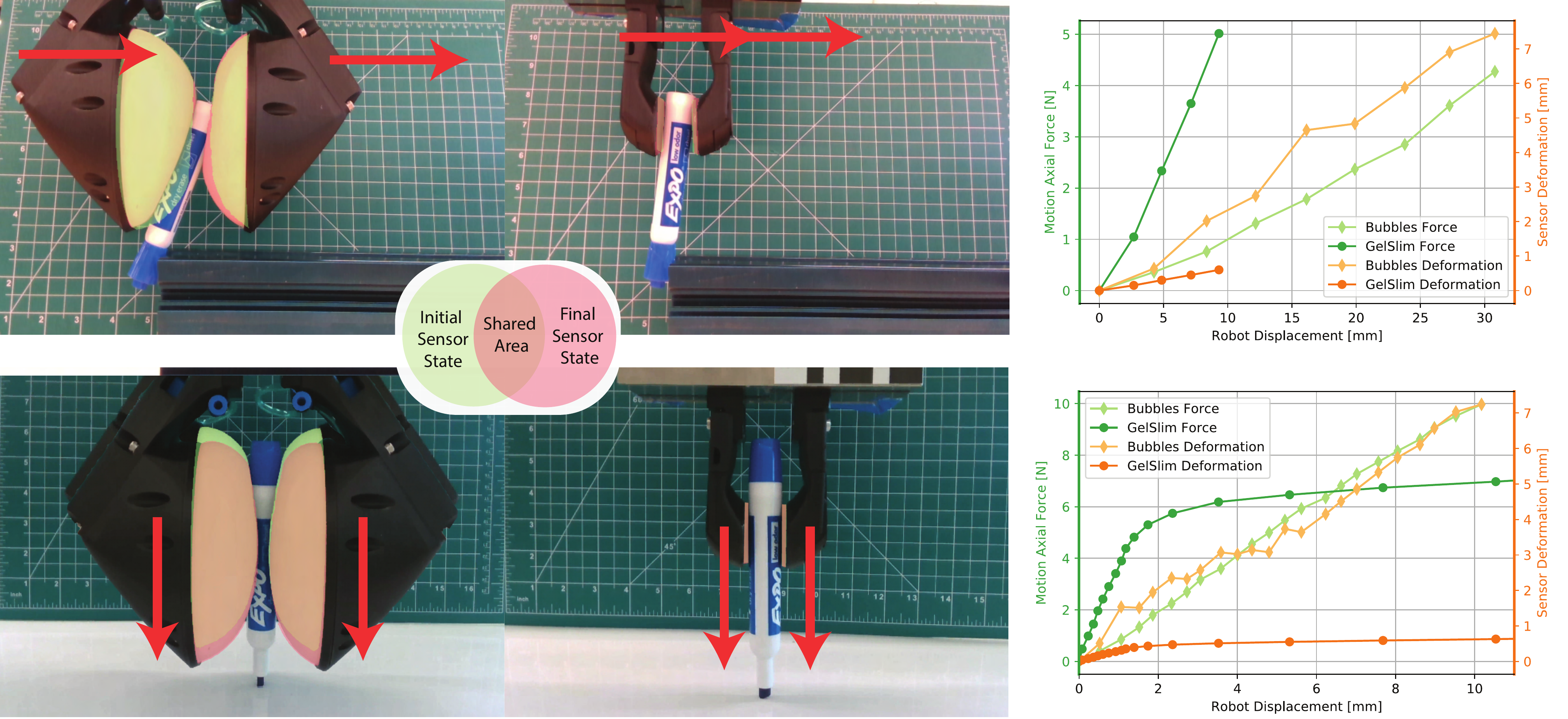}
    \caption{\textbf{Sensor Deformation Comparison} We compare the Soft Bubbles (left) and the GelSlim 3.0 (center) sensors on force transmission and deformation for two in-contact motions. Top row compares the sensors on an axial motion perpendicular to the grasp plane. Bottom row compares the sensor on a top-down axial motion in the grasp plane. Green overlay shows the sensor state before contact. Red overlay shows the deformed sensor state as a result of the interaction.}
    \label{fig:bubbles_vs_gelsight}
\end{figure}

Controlling both in-hand object pose and transmitted force is key for dexterous manipulation. In this section, we ask just how significant is the relative object displacement w.r.t. to the end-effector for soft vs hard sensors and how is transmitted force affected? Fig.~\ref{fig:bubbles_vs_gelsight} shows an illustrative comparison of the Soft Bubbles and Gelslim 3.0 for two in-contact motions along the sensor's main axes. The results show that the Soft Bubble membrane deforms an order of magnitude more than the GelSlim. We also observe that the object orientation varies significantly more (approx. $25^{\circ}$) during tangential motion. Moreover, we observe significant slip between the object and GelSlim, while the Bubbles maintain sticking contact due to large contact patches resulting from the significant deformation.

The GelSlim deformations are less than 1 mm and the resulting force transmission profile is relatively sharp, similar to rigid-body interactions. We also observe force plateauing and relative slip much earlier in the vertical contact task. In contrast, the soft sensors' large compliance allows for a more gradual force transmission without slip for a larger range of motion. While this compliance is desirable for many contact-rich tasks, it must be accounted for during fine manipulation.

The large contact patches result in increased surface area which provides two useful features: increased perception of the object shape at contact and larger distribution of friction through a concave contact surface. The former point can improve in-hand object pose estimation because it provides more features for inference, and the latter can improve grasp stability owing to larger and more distributed frictional forces where the membrane ``hugs’’ the object. The gradual build up of force is effectively a form of passive compliance. This compliance prevents damage to position controlled robots by mitigating rigid-on-rigid body contact. Further, it eases the burden on impedance control by providing local to contact compliance that can be perceived and controlled through robot motion. Thus, the remainder of this paper investigates how to learn the membrane dynamics of Soft Bubble sensors in order to realize the above benefits for dexterous manipulation.



\vspace{-5pt}
\subsection{Membrane Dynamics Model}
\label{sec:membrane_dyn} 
\vspace{-5pt}

\begin{figure}[b]
    \vspace*{-20pt}
    \centering
    \includegraphics[width=\textwidth]{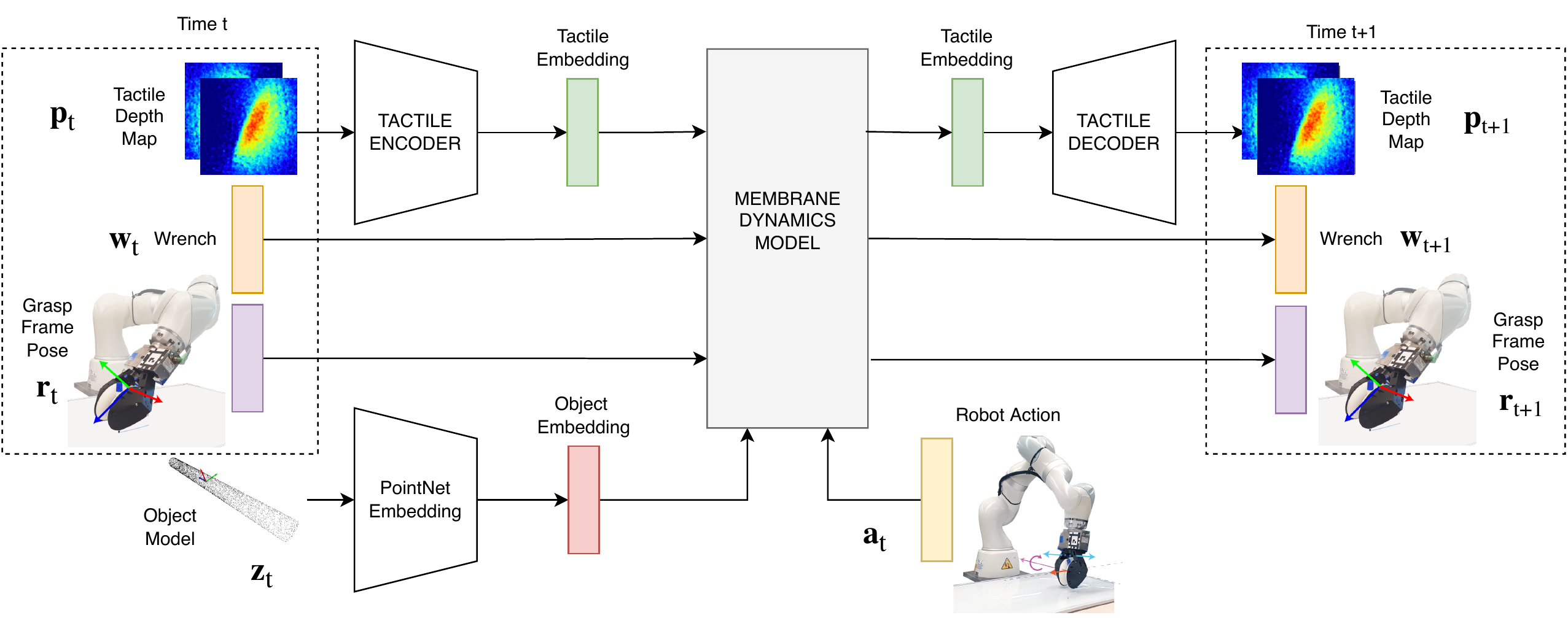}
    \caption{\textbf{Membrane Dynamics Model} Our proposed dynamics model predicts the membrane states $\vec{p}$, external wrenches $\vec{w}$ and grasp frame poses $\vec{r}$ conditioned on the robot action $\vec{a}$ and the grasped object model $\vec{z}$. The high-dimensional tactile depth maps and object models are projected into a learned lower-dimensional space for more efficient processing. Tactile embedding projections are learned in an autoencoder fashion. Object models are encoded using a PointNet-based network.} 
    \label{fig:dyn_model}
    \vspace*{-10pt}
\end{figure}

The goal of the membrane dynamics model is to predict the tactile sensor membrane deformations $\vec{p}_{t+1}$ and reaction forces $\vec{w}_{t+1}$ as a function of their current geometry $\vec{p}_{t}$, reaction force $\vec{w}_{t}$, the robot action $\vec{a}_{t}$, and the tool geometry $\vec{z}_{t}$.
The model has access to the current membrane geometry measured by a time-of-flight depth sensor mounted within the gripper. This membrane geometry $\vec{p}$ can be represented as either a depth map image or a 3D point cloud using the camera intrinsics. Here, we use the depth map encoding as it is a more structured and simplifies pixel-wise correspondences.

The dynamics model predicts the pixel-wise deformation of the sensor membrane and the reaction force sensed at the end-effector using the architecture depicted in Fig. \ref{fig:dyn_model}. In order to enable multi-step planning, the model also predicts the grasp frame pose reached by the robot $\vec{r}_{t+1}$ as a consequence of taking the action $\vec{a}_t$ (accounting for the robot impedance).
Similar to \cite{van2016stable}, the model encodes tactile images into a lower-dimensional latent representation. Intuitively, the model learns to exploit structure in the deformed membrane geometry to obtain a more compact representation. 
The object geometry is also encoded into a lower-dimensional representation to help the membrane dynamics model combine these multi-modal inputs. 
Since the object geometry is provided as a point cloud, our model uses a PointNet \cite{pointnet} inspired network pre-trained on ModelNet40 \cite{modelnet40} and fine-tuned on our task data to obtain a lower-dimensional latent representation of the grasped object.

We train our dynamics model in two-steps. First, we learn the tactile image latent representation in an Autoencoder \cite{rumelhart1985learning} fashion with only the tactile images from our training dataset. Our model is able to exploit tactile data from different tasks by training the tactile encoder-decoder on the combined dataset, subsequently sharing this latent tactile representation across tasks.
We also pre-train the object geometry model embedding on ModelNet40 and freeze all weights but the last layer. Second, we freeze the tactile encoder-decoder and train the dynamics model end-to-end.

We train the dynamics model $f$ with data composed by state-action-state transitions conditioned on the object geometry embedding $(\state_t, \vec{z}_t, \vec{a}_t, \state_{t+1})$.  In our case $\state_t = (\vec{p}_t, \vec{w}_t, \vec{r}_t)$.
We employ a supervised reconstruction loss $\mathcal L_\text{dyn}$ on the predicted state composed of 3 mean squared error (MSE) terms, one for each of the three sensory modalities our model predicts: tactile, wrenches, and poses. The losses are aggregated based on weights $\alpha_i$ to compensate for the discrepancies in units and scale of each sensory contribution (see Appendix \ref{supp:membrane_dynamics_impelentation} for implementation details).
\begin{align*}
    \mathcal L_\text{dyn}(\predvec{\state}_t, \vec{\state}_t) &= \alpha_1 \text{MSE}(\predvec{p}_{t+1},\vec{p}_{t+1}) + \alpha_2 \text{MSE}(\predvec{w}_{t+1},\vec{w}_{t+1}) + \alpha_3 \text{MSE}(\predvec{r}_{t+1},\vec{r}_{t+1}) \\ 
    &\text{where }\quad \predvec{p}_{t+1}, \predvec{w}_{t+1}, \predvec{r}_{t+1} = f( \vec{p}_{t}, \vec{w}_{t}, \vec{r}_{t}, \vec{z}_{t}, \vec{a}_{t})
\end{align*}


\subsection{Observation Model}

\begin{wrapfigure}[33]{r}{0.48\textwidth}

    \vspace*{-30pt}
    \centering
    \begin{subfigure}[b]{.48\textwidth}
    \centering
    \includegraphics[width=\textwidth]{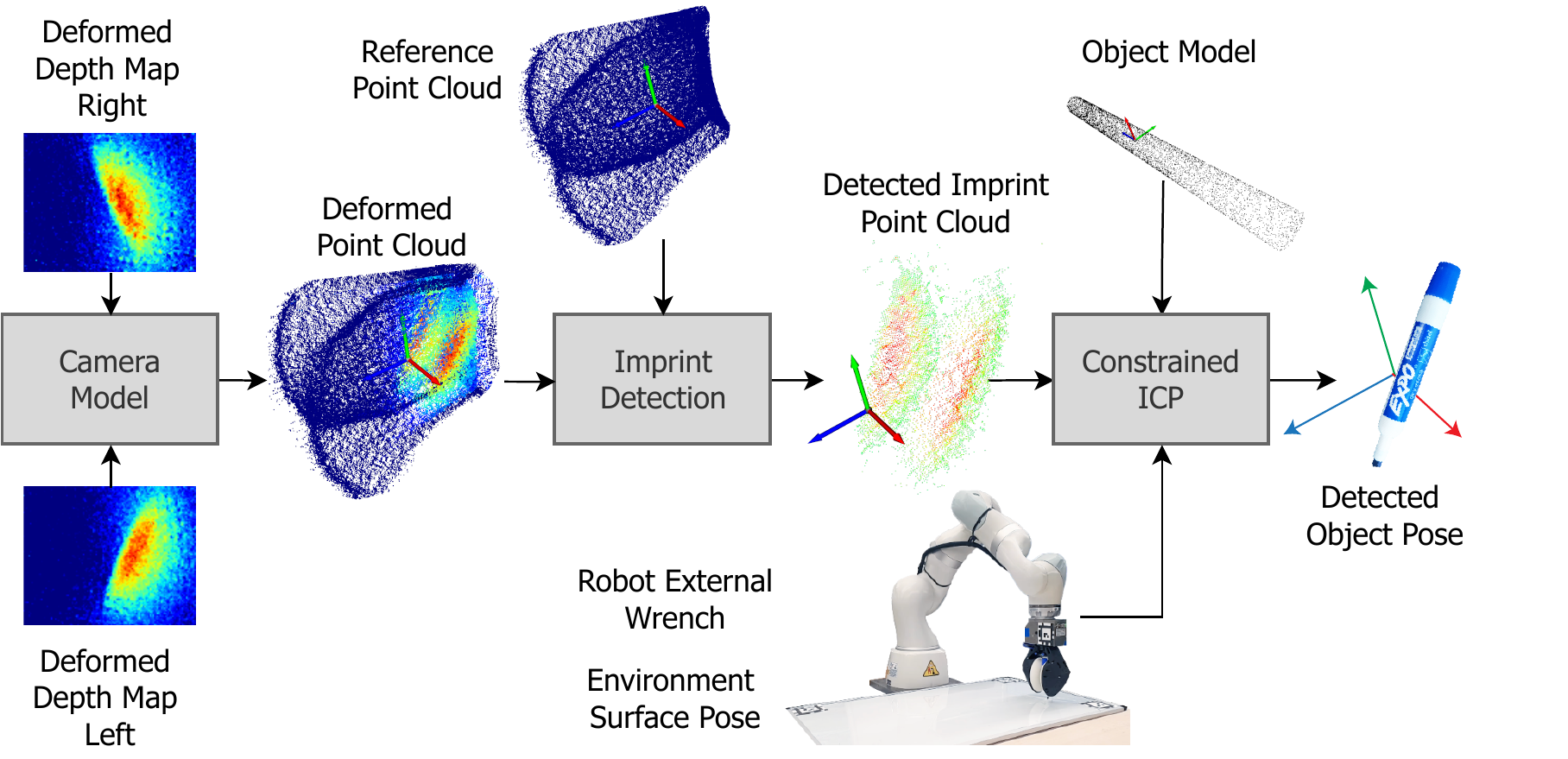}
    \caption{\textbf{Observation Model} The deformed depth maps are projected using the camera intrinsics and extrinsics to obtain the joint pointcloud of the deformed sensors. Then this is compared with a reference pointcloud of the undeformed sensor state to extract the contact points. Finally, ICP fits an object model to the detected contact points to estimates the corresponding object pose.}
    \label{fig:obs_model}
    \end{subfigure}
    
    \begin{subfigure}[b]{.48\textwidth}
    \centering
    \includegraphics[width=\textwidth]{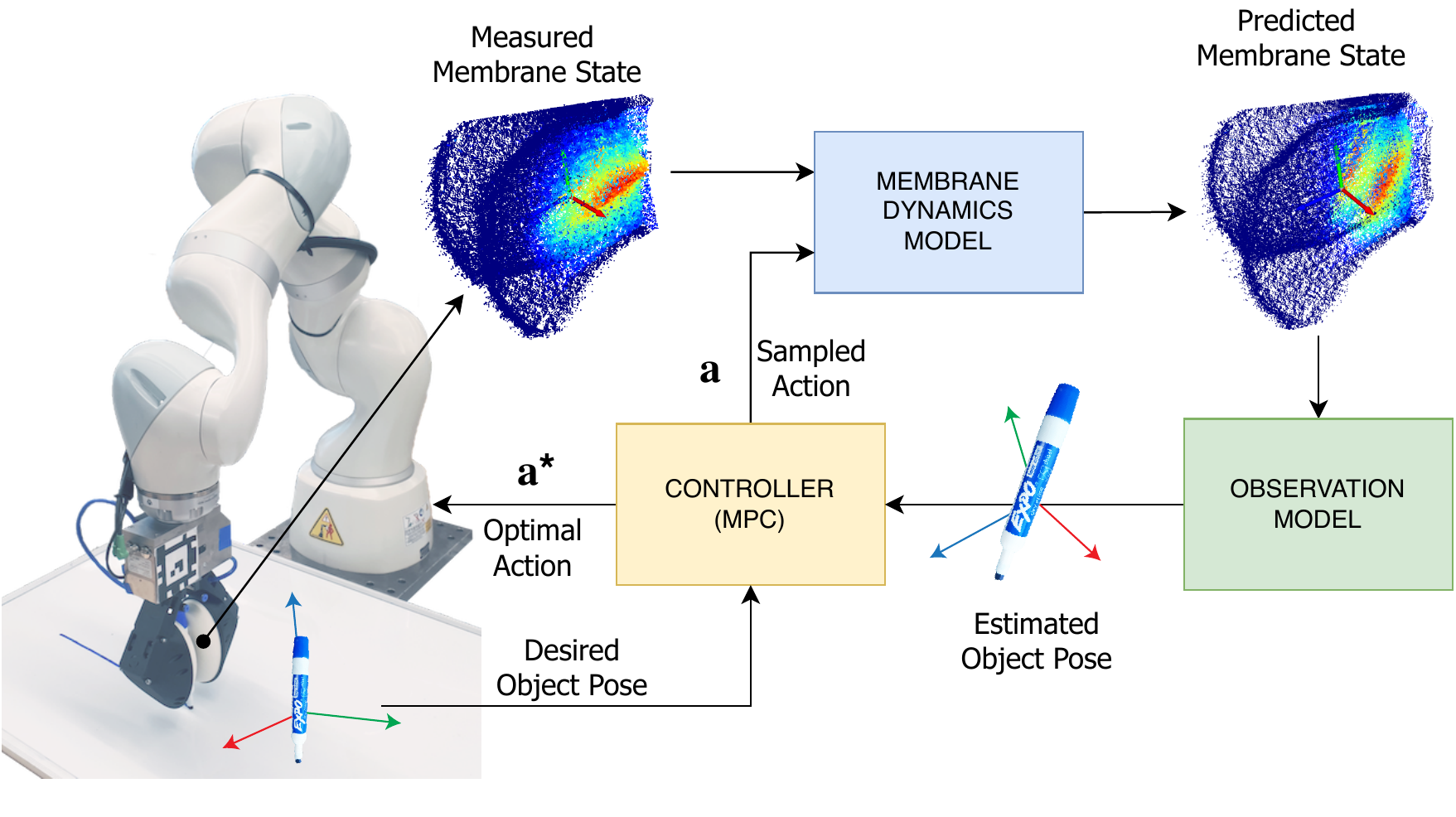}
    \caption{\textbf{Control Pipeline} Given a measured state, our controller queries the membrane dynamics model with sampled actions to obtain the predicted membrane states. The object pose is estimated from the predicted membrane states and it is compared with the desired poses to compute the costs associated to the sample actions. The costs are aggregated and the resultant optimal action is executed by the robot.}
    \label{fig:control_pipeline}
    \end{subfigure}
    \vspace*{-15pt}
    \caption{}
\end{wrapfigure}

The observation model $g$ maps the current membrane geometry $\vec{p}_t$ and end-effector reaction wrench $\vec{w}_t$ to an estimate of the current grasped object pose $\vec{q}_t$. It has access to the object geometry $\vec{z}_t$, robot proprioception $\vec{r}_t$, and the location of the environment contact surface.
We decouple pose estimation from membrane dynamics, enabling flexibility in observation model choice and improving generalization across objects and tasks.
Fig.~\ref{fig:obs_model} illustrates the observation model. First, we use the camera intrinsics and extrinsics to project the depth maps into the end-effector frame. Next, we extract contact points by comparing the current tactile images to their undeformed reference states. We extract contact points by computing pixel-wise deformation and selecting the top 10\% that deform at least 3 mm. See Appendix \ref{supp:observation_model} for more details. Next, we use Iterative Closest Points (ICP) to estimate the object pose from the extracted contact points. When the grasped object is in contact with the environment, we project the predicted object pose to the object-environment contact manifold \cite{koval2015pose}.

\vspace{-5pt}
\subsection{Controller}
\vspace{-5pt}

The goal of the controller is to drive the task states $\taskstate_t = (\vec{q}_t, \vec{w}_t)$, composed by object-environment configurations $\vec{q}$ and transmitted forces $\vec{w}$, to their goal values $\taskstate_g = (\vec{q}_g, \vec{w}_g)$ by solving the optimization problem posed in Sec.~\ref{sec:prob-form}. To this end, we use model-predictive control; specifically, the Model-Predictive Path Integral (MPPI) controller \cite{mppi1}. MPPI is effective in handling continuous action spaces and can be parallelizable for an efficient implementation with neural network models. Fig.~\ref{fig:control_pipeline} illustrates the control pipeline. The controller iteratively optimizes a nominal control sequence by sampling action sequences and rolling out their dynamics using our learned dynamics model and observation model.
These trajectories are sequences of task states $\taskstate_t, \dots, \taskstate_{t+N}$, and they are obtained by first using our learned dynamics model to obtain dynamic states $\dynstate_i$, and then using the observation model to obtain the correspondent task states $\taskstate_i$.
Finally, the trajectory costs are computed by comparing the predicted task states, with the desired values under a quadratic cost and the nominal control sequence is updated. 
For pivoting, the goal object orientations are defined as the desired tool orientation with respect to the robot. For drawing, the goal object orientations are perpendicular to the whiteboard along the drawing outline. In both cases the desired forces are perpendicular to the environment surface to encourage contact.
See Appendix \ref{supp:controller_implementation} for implementation details.


%% file: sections/Experiments.tex
\vspace{-5pt}
\section{Experiments and Results}\label{sec:experiments}
\vspace{-7pt}

\begin{wrapfigure}[19]{r}{0.48\textwidth} 
    \vspace*{-25pt}
    \centering
    \includegraphics[width=0.48\textwidth]{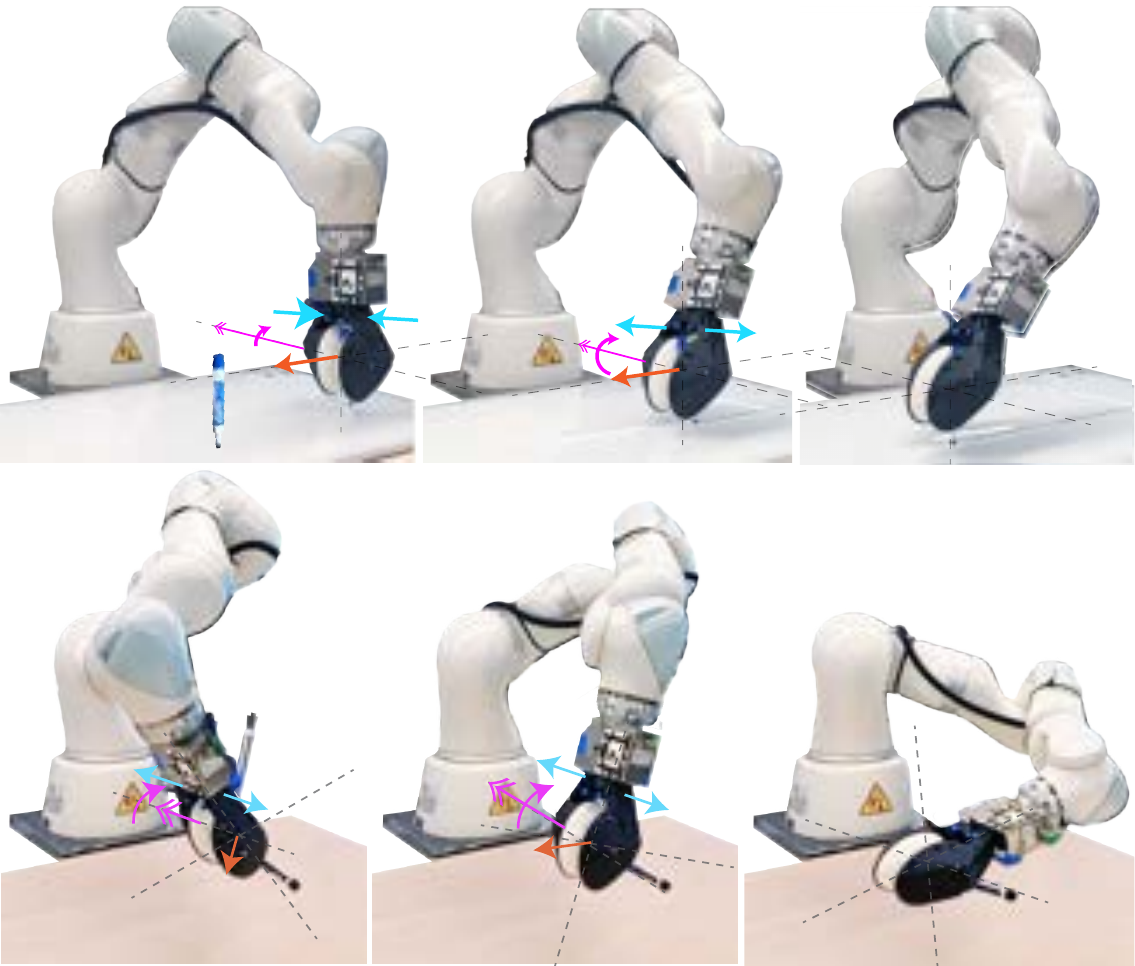} 
    \caption{\textbf{Action Spaces} The action spaces for drawing (top) and pivoting (bottom) are composed of the grasp width, displacement along the grasp plane, and rotation about axis perpendicular to this plane.}
    \label{fig:action_spaces}
\end{wrapfigure}
We demonstrate our proposed approach to tactile control in two real-world tasks: drawing with a grasped marker and in-hand pivoting of objects. Both of these tasks require the robot to apply forces through the grasped object to the environment while controlling its pose with respect to the end-effector. Since the grasp is not rigid, the object may move w.r.t. the end-effector due to deformations or slip. Effective task execution must account for relative motion and prevent failure due to the tool slipping out of the grasp.

\vspace*{3pt}
Our method uses the same dynamics model architecture (Fig. \ref{fig:dyn_model}) and learned latent membrane geometry representation across tasks. 
One key benefit of learning membrane deformations as opposed to object poses is that although each task has different dynamics, the membrane deformation dynamics are inherent to the sensor. This allows us to combine data from multiple tasks to learn the membrane latent representation.
However, since each task has particular dynamics and action spaces, the membrane dynamics function (Fig. \ref{fig:dyn_model} gray block) is task-specific.

\begin{figure}[]
    \centering 
    \includegraphics[width=\textwidth]{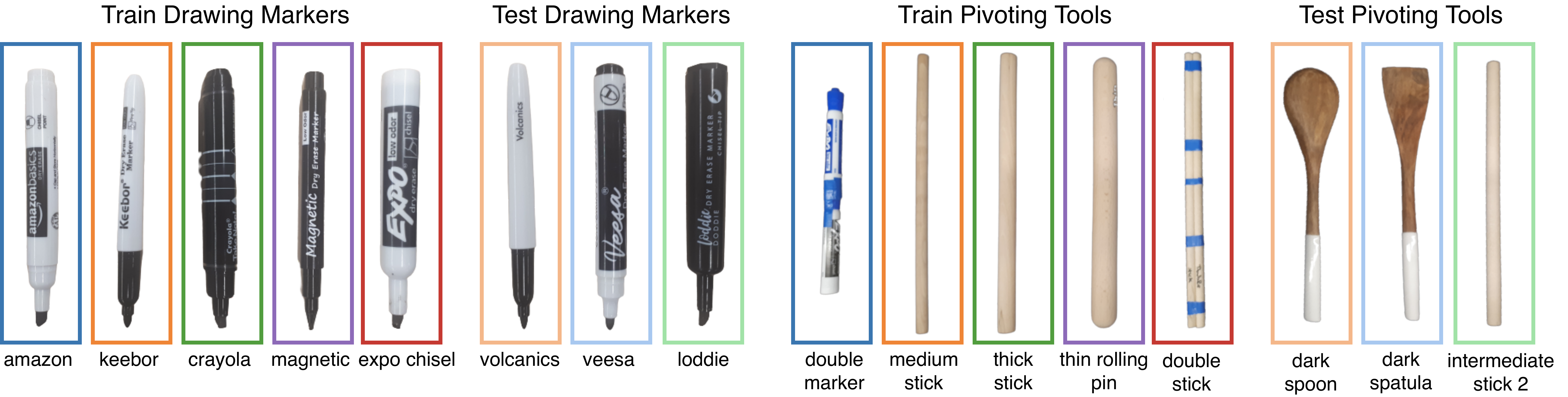}
    \vspace*{-17pt}
    \caption{\textbf{Evaluated Tools} (left) drawing object set, (right) tools used for pivoting. }
    \label{fig:task_tools}
    \vspace*{-17pt}
\end{figure}

We evaluate our two tasks on multiple tools, 8 for drawing and 8 for pivoting, Fig.~\ref{fig:task_tools}. For each task, we split our tools into two sets: we collect training data with 5 tools (train object set), and use the remaining 3 to test how well our learned model generalizes to unseen objects (test object set).
Using the 5 training tools, we collect 800 state-action-state triplets per tool, obtaining a total of 4000 samples to train the models.
Our data collection combines random samples with epsilon-greedy samples from a Jacobian controller (see Appendix \ref{supp:data_collection} for details). 

\vspace{-5pt}
\subsection{Baselines}
\vspace{-5pt}

We benchmark our proposed dynamics model against 5 baselines inspired by related approaches.
The original baselines only consider tactile signatures as inputs. Hence, we extend each to also incorporate reaction wrenches, robot poses, and object geometries.
We also modify the baselines such that all evaluated methods use the same observation model. This way, all methods have access to the same information for a fair comparison.
Our first baseline, \textit{Bubble Linear Dynamics} is inspired by \cite{van2016stable} where the dynamics are constrained to be linear in the tactile latent space.
Our second baseline, \textit{Object Pose Dynamics}, is inspired by \cite{lambeta2020digit} where the object in-hand pose dynamics are modeled as opposed to membrane dynamics. We adapt this baseline by replacing the object pose estimator with our observation model for consistency across all methods. 
Our third baseline, \textit{Fixed Model}, is only used in the drawing task. It assumes that there are no dynamics and that the membrane state remains constant. This baseline illustrates the importance of modeling membrane deformations. 
Our fourth baseline, \textit{Jacobian}, is only used in the pivoting task and is inspired by \cite{li2013control}. This baseline assumes that the grasped object is fixed w.r.t. the environment during contact.
Our final baseline does not use a dynamics model. Instead, the controller takes pseudo-random actions. For drawing, pseudo-random actions move the robot along the drawing path while uniformly randomly selecting the end-effector orientation, grasp width, and height from the action space. For pivoting, these actions bring the object into contact with the table while randomly selecting subsequent end-effector motion and grasp width. This baseline evaluates the complexity of each task.

\vspace{-5pt}
\subsection{Drawing with a Marker}
\vspace{-5pt}

The goal of this task is to control the marker pose and force transmitted through its tip to draw a desired shape while preventing the marker from slipping out of the grasp.
For simplicity, we choose a continuous 0.6 meter line. Fig.~\ref{fig:action_spaces} shows the robot action space composed of Cartesian displacement in the plane of the line, rotation about an axis perpendicular to this plane, and the gripper width. 

\begin{figure}[]
    \vspace{-15pt}
    \includegraphics[width=\linewidth]{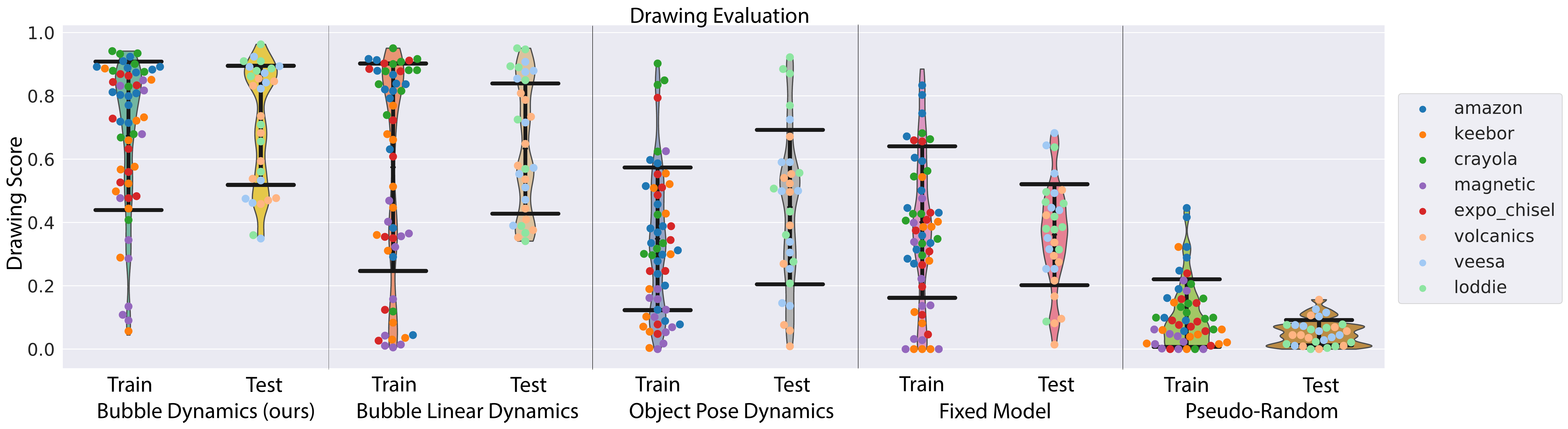}
    \label{fig:drawing_evaluation}
    \vspace*{-15pt}
    \caption{\textbf{Drawing Evaluation Results} 
    In pairs, we show the train (left) and test (right) scores for each of the evaluated methods. Colored dots indicate each tool achieved score (10 per tool). Black horizontal lines indicate the sample standard deviation around the mean. The drawing score represents the percent of the desired drawing successfully drawn, being 1.0 the optimal.
    }
    \label{fig:task_evaluation}
\end{figure}

We evaluate the drawing performance using a camera to extract a binary mask of the white board and comparing it to the desired binary image (see Appendix \ref{supp:drawing_evaluation} for more details). The drawing score is a value between 0 and 1, representing the percent of the line the robot has successfully drawn. The goal states provided to the controller are a sequence of poses where the marker tip is perpendicular to and in contact with the board along the desired drawing path. Fig.~\ref{fig:task_evaluation} shows the drawing score distributions. Table \ref{tab:evaluation_results} summarizes the drawing evaluation scores. We perform 10 trials per marker. The trials end when the final desired pose is reached, if the maker slips out of the hand, or if the robot can no longer correct the marker pose because of joint limits or collisions. 
Our results show that nonlinear membrane model (ours) outperforms the baselines both in training and test objects. For train objects, our model successfully draws on average a 10.0\% more of the desired line than the second best baseline (\textit{Bubble Linear Dynamics}), and a significant 32.6\% more than \textit{Object Pose Dynamics}. For test objects, the difference is 7.4\% and 25.8\% respectively.
We believe this is due to better robustness to uncertainty, generalization, and expressivity (see Sec.~\ref{sec:discussion}). 
Our results also show that \textit{Object Pose Dynamics} performs fairly similar to \textit{Fixed Model}.
Moreover, in general test objects perform slightly better on average than the training objects. A possible explanation is that their geometry and tip properties make the test objects easier to control, something we could not know \textit{a priori}. The ``magnetic'' brand marker is particularly challenging likely due to its fine tip. Removing this marker from the evaluation metrics eliminates the discrepancy between test and training sets, suggesting that the model has effectively learned important salient features of the task that it can indeed generalize to novel markers.



\vspace{-5pt}
\subsection{In-Hand Pivoting}
\vspace{-5pt}

The goal of this task is to drive the grasped object to a desired in-hand configuration solely using the environment to modify its configuration w.r.t the end-effector. Fig.~\ref{fig:action_spaces} shows the action space of the pivoting task which is composed of Cartesian motion in the plane of the gripper jaws, rotation about an axis perpendicular to this plane, and the gripper width.
Our metric for pivoting performance is error in desired vs. realized tool orientation. We use our observation model to estimate the final pose. The desired tool pose is sampled at random within the graspable range. The controller optimizes the pivoting action sequence to achieve the final pose by exploiting the dynamics model. Table \ref{tab:evaluation_results} summarizes the pivoting evaluation scores (see Appendix \ref{supp:pivoting_results} for the evaluation distributions). We perform 10 pivoting executions per tool. The trials end when reaching the final desired pose within, or after having performed 10 pivoting actions. If the tool slips out of hand, we record the last achieved angle.
Our results show that our proposed model outperforms the baselines in reaching desired in-hand orientations, obtaining the lowest orientation error on average. Our method is also the most consistent, obtaining the lowest standard deviation. We highlight that our method obtains an orientation error as low as half the second best approach. Our results also show our method is less prone to overshooting errors compared to baselines.

\begin{table}[h!]
\vspace{-5pt}
\centering
\begin{tabular}{lcccccccc}
\toprule
\multicolumn{1}{l}{\multirow{3}{*}{Representation}} & \multicolumn{4}{c}{Drawing Scores } & \multicolumn{4}{c}{Pivoting Scores [deg]} \\ \cmidrule(lr){2-5}\cmidrule(lr){6-9}
\multicolumn{1}{c}{} & \multicolumn{2}{c}{Train Objects} & \multicolumn{2}{c}{Test Objects} & \multicolumn{2}{c}{Train Objects} & \multicolumn{2}{c}{Test Objects} \\ 
\multicolumn{1}{c}{} & \multicolumn{1}{c}{Mean $\uparrow$} & \multicolumn{1}{c}{Std $\downarrow$} & \multicolumn{1}{c}{Mean $\uparrow$} & Std $\downarrow$ & \multicolumn{1}{c}{Mean$\downarrow$} & \multicolumn{1}{c}{Std$\downarrow$} & \multicolumn{1}{c}{Mean$\downarrow$} & Std $\downarrow$\\ \midrule
\begin{tabular}[c]{@{}l@{}}Bubble Dynamics\end{tabular} & \multicolumn{1}{c}{\textbf{0.674}} & \multicolumn{1}{c}{0.235} & \multicolumn{1}{c}{\textbf{0.707}} & 0.188 & \multicolumn{1}{c}{\textbf{6.96}} & \multicolumn{1}{c}{12.2} & \multicolumn{1}{c}{\textbf{5.41}} & 6.33 \\ 
\begin{tabular}[c]{@{}l@{}}Bubble Linear Dyn.\end{tabular} & \multicolumn{1}{c}{0.574} & \multicolumn{1}{c}{0.328} & \multicolumn{1}{c}{0.633} & 0.206 & \multicolumn{1}{c}{14.4} & \multicolumn{1}{c}{19.8} & \multicolumn{1}{c}{11.5} & 12.8 \\ 
\begin{tabular}[c]{@{}l@{}}Object Dynamics\end{tabular} & \multicolumn{1}{c}{0.348} & \multicolumn{1}{c}{0.225} & \multicolumn{1}{c}{0.449} & 0.244 & \multicolumn{1}{c}{24.2} & \multicolumn{1}{c}{38.9} & \multicolumn{1}{c}{16.8} & 26.9 \\ 
\begin{tabular}[c]{@{}l@{}}Fixed / Jacobian\end{tabular} & \multicolumn{1}{c}{0.402} & \multicolumn{1}{c}{0.239} & \multicolumn{1}{c}{0.361} & 0.159 & \multicolumn{1}{c}{22.8} & \multicolumn{1}{c}{36.7} & \multicolumn{1}{c}{9.39} & 20.5 \\ 
\begin{tabular}[c]{@{}l@{}}Pseudo-Random \end{tabular} & \multicolumn{1}{c}{0.114} & \multicolumn{1}{c}{0.107} & \multicolumn{1}{c}{0.050} & 0.040 & \multicolumn{1}{c}{16.9} & \multicolumn{1}{c}{16.9} & \multicolumn{1}{c}{24.6} & 37.4 \\ \bottomrule
\end{tabular}
\vspace*{3pt}
\caption{\textbf{Evaluation \& and Task Score Statistics:} Drawing scores = percentage of the drawing the robot has successfully completed represented as a number between 0 and 1 (higher = better). Pivoting scores = absolute value of the error in desired vs goal orientation. (lower = better). Statistics are reported over 10 trials per representation.}
\label{tab:evaluation_results}
\end{table}

%% file: sections/Discussion.tex
\section{Discussion and Limitations}\label{sec:discussion}

In this paper, our experiments show that decoupling membrane dynamics from object dynamics and characterizing the sensor dynamics improves task performance as well as the ability to use novel unseen tools.
We believe that modelling sensor dynamics performs better than object dynamics because: i) sensor dynamics explicitly considers the combined compliance of the robot and the sensor, ii) are more robust against observation noise, and iii) can share tactile signatures across tasks and objects. Our observation model has some stochasticity, and its variance can highly impact trajectories that roll out from its observations. Instead, our approach rolls out dynamic trajectories agnostic to the observation model. As a result, the final pose prediction is less effected by the observation variability. See Appendix \ref{supp:membrane_vs_object_pose} for a detailed comparison.


A limitation of our approach is that it assumes rigid objects with known geometry. This assumption simplifies the observation model, planning, and controls, since it restricts object-environment interactions to rigid on rigid. While the membrane dynamics model may not be affected significantly by the object compliance, our method will require a more sophisticated observation and controls framework.
Another limitation of our model is the quality of long-horizon predictions. Long-horizon predictions can be poor because recursive model calls may lead to out of distribution predictions. This is the main reason why we use a planning horizon of 2 steps. 
Some methods to mitigate this effect are multi-step prediction methods \cite{hirose2019deep} or flow-based projections \cite{power2022variational}. Another direction to explore is developing new membrane representations that are robust to this effect.
Finally, our model predictions are smooth. As a consequence, our method may fail to model drastic state changes like the outcomes of hitting obstacles, or dropping the tool. This could be addressed by leveraging the structure of non-smooth/hybrid dynamics \cite{pfrommer2020contactnets} during representation learning.


%% file: sections/Supplementary.tex
\newpage
\begin{appendices}
\counterwithin{figure}{section}

\section{Implementation Details}
\label{supp:implementation_details}
\subsection{Sensor Evaluation Details}
\label{supp:sensor_eval}
We evaluate two state-of-the-art dense tactile sensors such as the GelSlim 3.0 \cite{taylor2021gelslim3} and the Soft Bubbles \cite{alspach2019soft}. We compare the sensors in terms of deformations and the transmitted forces. We evaluate the sensors' compliance since it produces tool motion with respect to the robot's end-effector.

Both sensors have dot patterns imprinted on their contact surfaces to enable surface tracking and deformation estimation.
The imprinted dots can be detected using segmentation methods. We use Gaussian-weighted adaptive thresholding to binarize the sensor images and then find the dot contours.
We use Gunnar-Farneback Dense Optical flow method \cite{farneback2003two} to track the dots motions, extract the deformation fields, and compute correspondences. 
We project the dots deformations into their spatial coordinates for the reference and deformed state and use the correspondences to compute each dot's deformation.
To estimate the deformation perpendicular to the camera plane, we use the measured depth via their depth camera for the Bubbles sensors. For the GelSlims, we exploit the known rigid object geometry.

We compare the sensors response on two axial motions: Axial motion perpendicular to the grasp frame (Fig. \ref{fig:bubbles_vs_gelsight} top) and a top-down motion along the sensor's contact plane main axis (Fig. \ref{fig:bubbles_vs_gelsight} bottom). 
The robot is controlled in Cartesian impedance mode and commanded to move in increments of 5mm.
We measure steady-state states. We use the robot to extract the motion axial force. For the deformation, we report the average of the 5\% of the dots that deform the most.

\subsection{Membrane Dynamics Model Implementation}
\label{supp:membrane_dynamics_impelentation}

We implement our membrane dynamics model in PyTorch. We use PyTorch Lightning for efficient and easy training and logging.

\subsubsection{Architecture}
\begin{figure}[H]
    \centering
    \includegraphics[width=\textwidth]{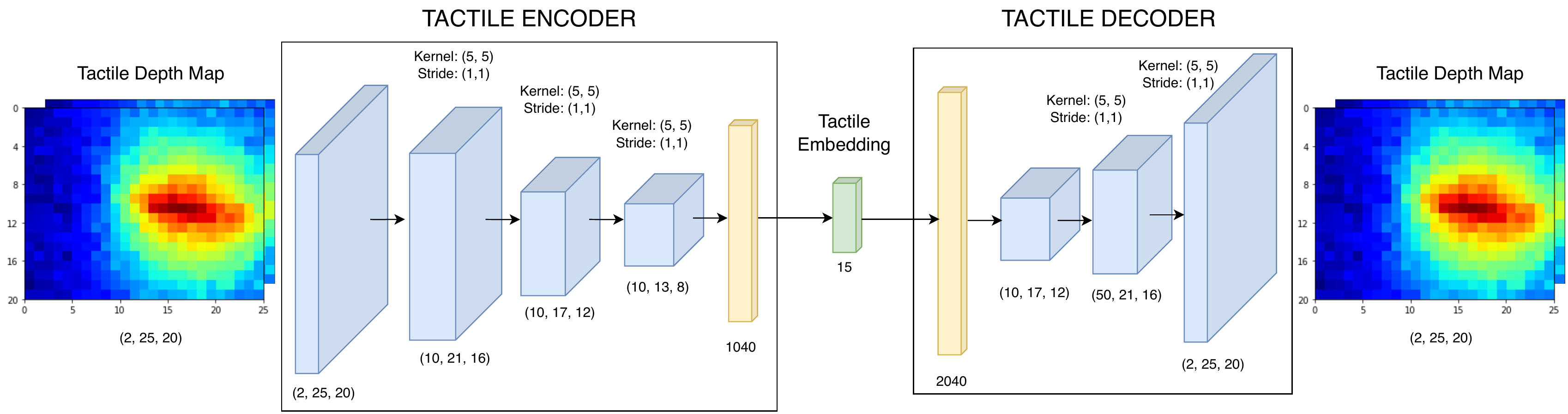}
    \caption{\textbf{Tactile Projection Architecture}  An encoder (left) maps the combined tactile depth maps for left and right bubbles into a 15-dimensional embedding. The decoder (right) reconstructs the tactile depth map from the embedding.} 
    \label{fig:tactile_embedding}
\end{figure}

Figure \ref{fig:dyn_model} illustrates our dynamics model. It is composed by 4 main elements:
\begin{itemize}
\item \textbf{Tactile Encoder}: The tactile encoder (Fig. \ref{fig:tactile_embedding} left) embeds a $(2\times 25\times 20)$ image into a 15-dimensional vector. It is composed by 3 Convolutional Neural Network (CNN) with $(5\times 5)$ kernels followed by 1 linear layer. The tactile encoder uses ReLU activations and batch normalizations. 
\item \textbf{Tactile Decoder}: The tactile decoder (Fig. \ref{fig:tactile_embedding} right) attempts to invert the encoder. It is composed by a linear layer followed by 2 deconvolution layers. It also has ReLU activations and batch normalizations.
\item \textbf{Membrane Dynamics Model}: As Fig. \ref{fig:dyn_architecture_comparison} A shows, the membrane dynamics model is a 3-layer neural network with ReLU activations and hidden sizes of 200. Its input is a concatenation of the tactile embedding (15-dimensional), wrench (6-dimensional), grasp frame pose (6-dimensional), object embedding (10-dimensional), and robot action (4-dimensional). The grasp frame pose is represented as a position and axis-angle orientation. It predicts the next tactile embedding (15-dimensional), next wrench (6-dimensional), and the correction of the grasp frame pose (6-dimensional).
\item \textbf{Object Embedding}: This is a PointNet \cite{pointnet} classifier network with last layer replaced by a 10-dimensional linear layer with no output activation to produce the desired embedding. 
\end{itemize}

\subsubsection{Training Details}
\label{supp:training_details}
Given a dataset $\mathcal D = \{(\vec{s}_t, \vec{z}_t, \vec{a}_t, \vec{s}_{t+1})_i\}_{i=1}^{N}$, we perform a 80-20 train-validation split. 
We train our models using the train data until convergence and use the remaining validation data to asses the model performance to unseen data. We use model weights with the lower validation loss.
We train our dynamics model in a two-step process. First we train the tactile embedding, using training data from multiple tasks.
For the tactile embedding, we use a MSE reconstruction loss on the tactile depth maps.
\begin{align*}
    \mathcal L_\text{tactile}(\predvec{p}_t, \vec{p}_t) &= \text{MSE}(\predvec{p}_{t+1},\vec{p}_{t+1})\\ 
    &\text{where }\quad \predvec{p}_{t+1} = f_\text{decode}(f_\text{encode}(\vec{p}_t))
\end{align*}

Once we have trained the tactile embedding, we freeze it and learn the membrane dynamics for each specific task. We use Adam optimization with a supervised loss on the predicted state composed of 3 MSE terms, one for each of the three sensory modalities our model predicts: tactile, wrenches, and poses. The losses are aggregated based on weights $\alpha_i$ to compensate for the discrepancies in units and scale of each sensory contribution. In our case:

\begin{align*}
    \mathcal L_\text{dyn}(\predvec{s}_t, \vec{s}_t) &= \alpha_1 \text{MSE}(\predvec{p}_{t+1},\vec{p}_{t+1}) + \alpha_2 \text{MSE}(\predvec{w}_{t+1},\vec{w}_{t+1}) + \alpha_3 \text{MSE}(\predvec{r}_{t+1},\vec{r}_{t+1}) \\ 
    &\text{where }\quad \predvec{p}_{t+1}, \predvec{w}_{t+1}, \predvec{r}_{t+1} = f( \vec{p}_{t}, \vec{w}_{t}, \vec{r}_{t}, \vec{z}_{t}, \vec{a}_{t})
\end{align*}

In practice, we found that our model predictions obtained best results with $\alpha_3 = 0$. Since our robot impedance is considerably stiff, the robot motions can be well approximated to be rigid. With this consideration, the only compliance in the system is the sensors' compliance.

\begin{table}[h!]
    \centering
    \begin{tabular}{c c c}
        Parameter Name & Parameter Symbol & Value \\
        \hline
        \vspace*{-7pt}\\
        tactile loss weight & $\alpha_1$ & 1 \\
        wrench loss weight & $\alpha_2$ & 0.0001 \\
        robot pose loss weight & $\alpha_3$ & 0 \\
        learning rate & $\lambda$ & 0.001 \\
        betas & $(\beta_1, \beta_2)$ & $(0.9, 0.999)$\\
    \end{tabular}
    \vspace*{5pt}
    \caption{\textbf{Dynamics Model Hyper-parameters:} Hyper-parameters used for training the dynamics model.}
    \label{tab:my_label}
\end{table}

\subsection{Observation Model Implementation}
\label{supp:observation_model}

The observation model is the responsible to estimate the pose of the grasped object from the membrane state. Since we have a model of the object geometry, we use Iterative Closest Points (ICP) to estimate the best object pose from the membrane state. In particular, we use point-to-point ICP since both the membrane and the object geometry are collections of points.

First, we extract the membrane points that also belong to the object. We refer to such collection of points as imprint. To this end, we filter out all membrane points based on their deformation values. We select the top 10\% of the deformations that deform at least 3mm. 
We also filter out points that are more than 15mm to the imprint point cluster, since they are noise. 

Once we have extracted a clean imprint, we apply ICP to iteratively find the best transformation of the object model that minimizing the L2 distance with the imprint.  We initialize the translation estimate as the mean of the imprint points. We also initialize the orientation to be randomly within $20^\circ$ of the imprint main axis. We perform 20 ICP iterations.
The resulted estimated tool pose is expressed with respect to the grasp frame, which is also the frame that the membrane points are registered.
When in contact, we impose non-penetration to the estimated object pose and we project it to the contact manifold. Intuitively, the object pose is translated to have a shared point with the environment without compromising the ICP score. 
We detect contact based on the external wrench feedback. When the external wrench perpendicular to the environment surface normal is above a threshold, then the robot assumes that it is in contact with the environment. Typically, we set the threshold to be 1.5N.

Our observation model implementation is able to estimate multiple object poses for multiple membrane states in parallel. We have implemented the imprint detection and the ICP algorithm in PyTorch for fast and easy integration with the controller algorithm. This allows our algorithm to execute all operations in the GPU.

\subsection{Controller Implementation}
\label{supp:controller_implementation}

Since our dynamics models are PyTorch models, our controller implementation is based on \texttt{pytorch\_mppi} \cite{pytorch_mppi}. This package exploits batched operations for fast and efficient real-time implementation. Dynamic model queries, cost computations, and sampling are done on a GPU.
At each step, the controller samples action sequences around the nominal action sequence based on Gaussian noise. We initialize the nominal action sequence as the mean of the action space.
Then, the dynamics model is queried and the membrane states are obtained along the sampled trajectories. 
Next, the object poses and transmitted forces for each trajectory are estimated using the observation model. 
The trajectory costs are computed comparing the predicted task states, composed by the object pose and transmitted force, with the desired ones. Finally, the nominal action sequence is updated based on an importance sum where the weight is computed from the trajectory cost. Intuitively, the lower the cost, i.e. the closer the state to the desire one, the higher the contribution to the control sequence update. The optimal action is sent to the robot and then it is discarded. 

Table \ref{tab:controller_times} reports average computation times for our controller implementation during task execution. Our experiments are executed in a system equipped with a NVIDIA GeForce RTX 3070 GPU and an AMD Ryzen 9 3900XT CPU.

The individual state cost is formulated in the task-state $\vec{x} = (\vec{q}, \vec{w})$ and its value is computed as:
\begin{align*}
    \mathcal J(\vec{x}_t, \vec{a}_t) &= (\taskstate_t-\taskstate_g)^{T} \mat{Q} (\taskstate_t-\taskstate_g)+ \vec{a}_t^{T} \mat{R} \vec{a}_t\\
    &\text{where} \quad \mat{Q} = \mat{I}, \quad \mat{R} = \mat{0} \\
\end{align*}

Considering the different magnitudes and dimensions in $\vec{x}_t = (\vec{q}_t, \vec{w}_t)$, we further reduce the above expression by splitting it into 2 terms as shown below. We set $\alpha_w = 0.0001$.

\begin{equation*}
    \mathcal J(\vec{x}_t) = \mathcal L_\text{obj pose}(\bold q_t, \bold q_g) + \alpha_w \text{MSE}(\vec{w_t}, \vec{w}_g)
\end{equation*}

For the object pose cost, we combine position and orientation terms in $\vec{q}$ by exploiting the known object geometry $\vec{z}$. Since the object geometry is a pointcloud, $\vec{z} = \{\vec{z}_1, \dots, \vec{z}_M\} \ \text{where} \  \vec{z}_i \in \R^3$, we can transform each individual point with the $SE(3)$ transformation given by $\vec{q}$ and evaluate the distance between the point pose given by $\vec{q}_t$ and the goal pose $\vec{q}_g$ using the MSE metric.
\begin{equation}
    \begin{split}
    \mathcal L_\text{obj pose}(\bold q_t, \bold q_g) &= \frac{1}{M}\sum_{i=1}^{M} \left[\mat{H}(\vec{q}_t) \tilde{\vec{z}}_i - \mat{H}(\vec{q}_g)\tilde{\vec{z}}_i\right]^T\left[\mat{H}(\vec{q}_t) \tilde{\vec{z}}_i - \mat{H}(\vec{q}_g)\tilde{\vec{z}}_i\right] - 1 \\
    &= \frac{1}{M}\sum_{i=1}^{M}  \tilde{\vec{z}}_i^T \left[\mat{H}(\vec{q}_t)- \mat{H}(\vec{q}_g)\right]^T\left[\mat{H}(\vec{q}_t) - \mat{H}(\vec{q}_g)\right]\tilde{\vec{z}}_i - 1
    \end{split}
\label{eq:object_pose_loss}
\end{equation}

Where $\mat{H}(\vec{q}) \in SE(3)$ is the homogeneous transformation matrix associated to the pose $\vec{q}$ and $\tilde{\vec{z}}_i$ is the homogeneous vector $\tilde{\vec{z}}_i = \begin{bmatrix}\vec{z}_i^T & 1\end{bmatrix}^T$.

\begin{table}[h!]
    \centering
    \begin{tabular}{c c c}
        Parameter Name & Parameter Symbol & Value \\
        \hline
        \vspace*{-7pt}\\
        lambda & $\lambda$ & 0.01 \\
        horizon & $T$ & 2\\
        number of samples & $N$ & 100\\
        noise sigma & $\sigma_\text{noise}$ & $30\%$\\
    \end{tabular}
    \vspace*{5pt}
    \caption{\textbf{Controller Hyper-parameters:} MPPI hyper-parameters used by our implementation.}
    \label{tab:controller_hyper}
\end{table}

\begin{table}[h!]
    \centering
    \begin{tabular}{c c c}
        Process Name & Computation Time [s]\\
        \hline
        \vspace*{-7pt}\\
        Dynamic Model Queries &  $4\cdot 10^{-5}$ \\
        Pose Estimation &  0.6 \\
        Cost Computation & 0.4 \\
        Total Control Step & 3.0\\
    \end{tabular}
    \vspace*{5pt}
    \caption{\textbf{Computation Performance:} Time costs in seconds for our control implementation during task execution. See table \ref{tab:controller_hyper} for the evaluated hyper-parameters.}
    \label{tab:controller_times}
\end{table}

\subsection{Baselines Details}
\label{supp:baselines}
\begin{figure}[h!]
    \centering
    \includegraphics[width=\textwidth]{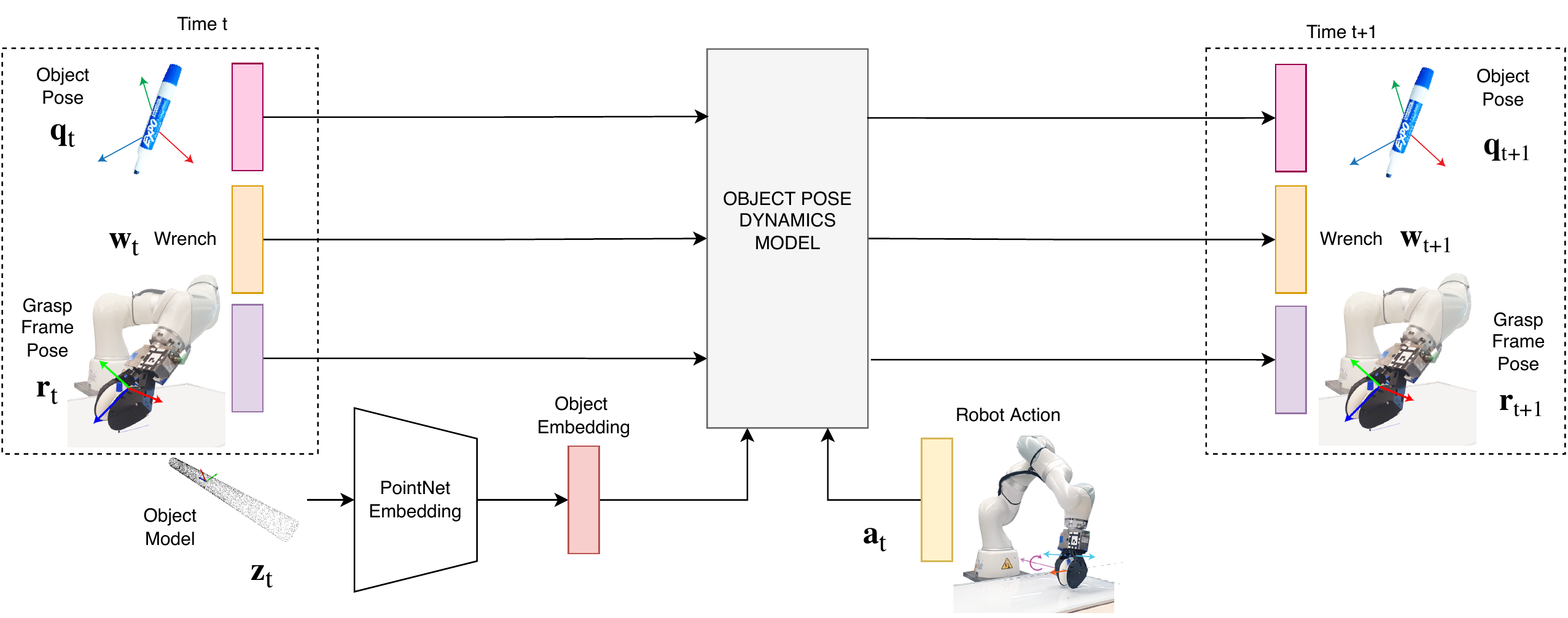}
    \caption{\textbf{Object Pose Dynamics} This architecture replace the tactile depth map for the object pose. As membrane dynamics, object model is still embedded using a PointNet network. However, since here there is not tactile data, we do not need the tactile encoder-decoder networks.}  
    \label{fig:obj_dyn_model}
\end{figure}

\begin{figure}[h!]
    \centering
    \includegraphics[width=\textwidth]{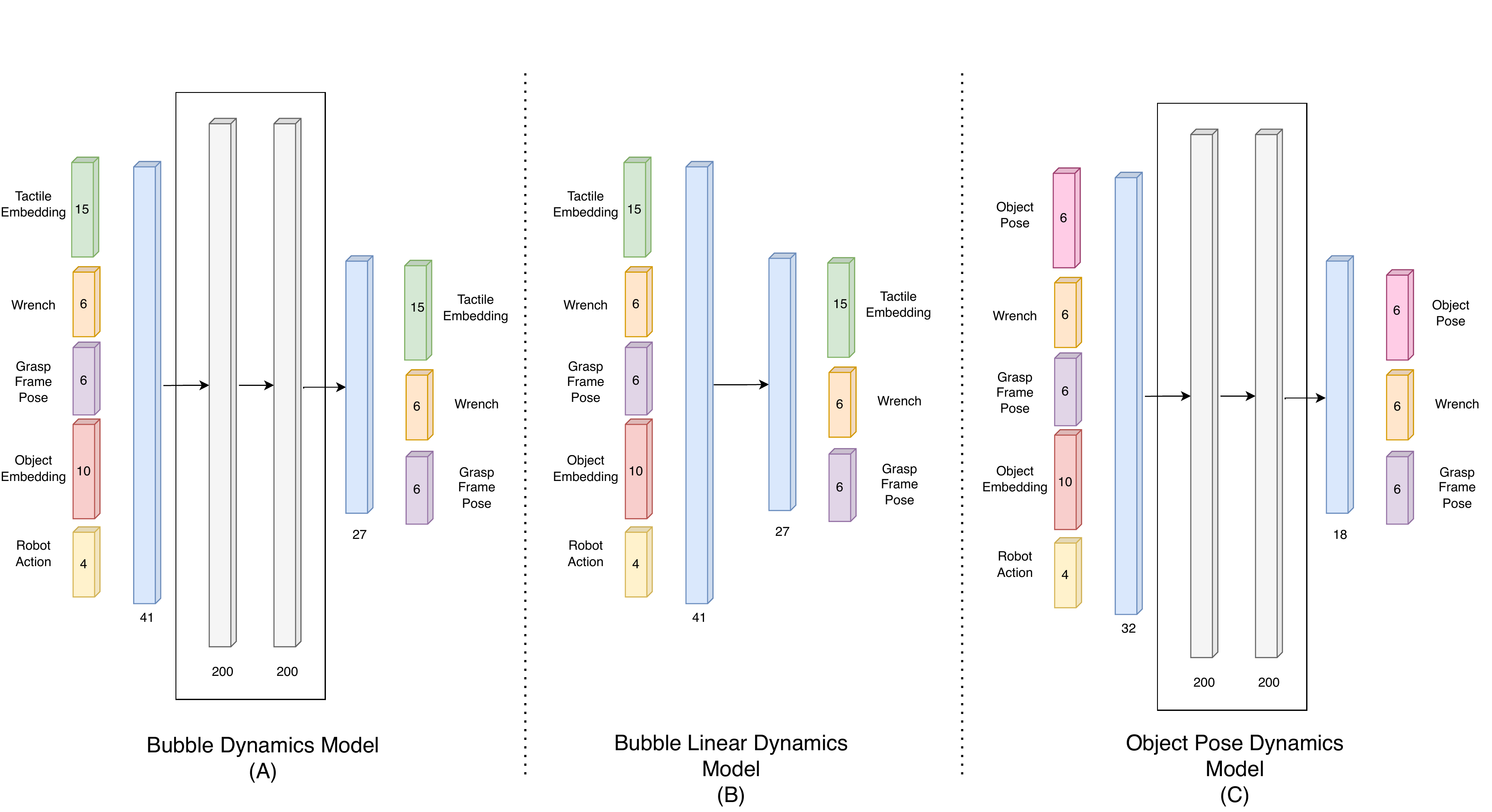}
    \caption{\textbf{Dynamics Architecture Comparison} Bubble Dynamics model (A) and Object Pose Dynamics Model (C) are 3-layer neural networks. They both have 2 hidden-layers with 200 units. Bubble Linear Dynamics (B) predicts the output based on a single linear layer with no biases. (A) and (B) use tactile embeddings, while (C) replaces them with object poses.}  
    \label{fig:dyn_architecture_comparison}
\end{figure}

\begin{itemize}
\item \textbf{Bubble Linear Dynamics}: This baseline constrains the membrane pose dynamics so next states are obtained with a linear operation with no biases. (Fig. \ref{fig:dyn_architecture_comparison} B). Therefore, the dynamics are:
\begin{equation*}
    \begin{bmatrix} \vec{p}_{t+1}^\text{emb}\\\vec{w}_{t+1}\end{bmatrix} = \mat{A}_\text{dyn} \begin{bmatrix} \vec{p}_t^\text{emb}\\\vec{w}_t\\\vec{r}_t\\\vec{z}_t\\ \vec{a}_t\end{bmatrix} \quad \text{where}\ \mat{A}_\text{dyn} \in \R^{21\times 41}, \quad \vec{p}_t^\text{emb}= f_\text{encode}(\vec{p}_t), \quad  \vec{p}_{t+1}= f_\text{decode}(\vec{p}_{t+1}^\text{emb})
\end{equation*}
\item \textbf{Object Pose Dynamics}: This baseline considers the object pose instead of the tactile signature (Fig. \ref{fig:obj_dyn_model}). Therefore, the state in this case is $\vec{s}_{\text{obj},t} = (\vec{q}_t, \vec{w}_t, \vec{r}_t)$. The object pose dynamics are modelled similar to the \textit{Bubble Dynamics} with a 2-hidden layers with 200 units (Fig. \ref{fig:dyn_architecture_comparison} C). Also, similar to the membrane model training, here we train the model with a loss composed by three term, but instead on a MSE over the tactile signatures, we have a loss term on the object pose. This object pose loss is defined in equation \ref{eq:object_pose_loss}.
\begin{align*}
    \mathcal L_\text{obj dyn}(\predvec{s}_{\text{obj}, t}, \vec{s}_{\text{obj}, t}) &= \alpha_1 \mathcal L_\text{obj pose}(\vec{q}_t, \predvec{q}_t) + \alpha_2 \text{MSE}(\predvec{w}_{t+1},\vec{w}_{t+1}) + \alpha_3 \text{MSE}(\predvec{r}_{t+1},\vec{r}_{t+1}) \\ 
    &\text{where }\quad \predvec{q}_{t+1}, \predvec{w}_{t+1}, \predvec{r}_{t+1} = f_\text{obj dyn}( \vec{q}_{t}, \vec{w}_{t}, \vec{r}_{t}, \vec{z}_{t}, \vec{a}_{t})
\end{align*}

\item \textbf{Fixed Model}: This baseline assumes that the membrane state remains constant ($ \vec{p}_{t+1} = \vec{p}_t$), i.e. there is no relative motion between the object and the robot. It also assumes that there is no change in the external sensed forces ($\vec{w}_{t+1} = \vec{w}_t$). Future grasp frame poses $\vec{r}_{t+1}$ are predicted using a deterministic robot action model considering the robot geometry, the current grasp pose $\vec{r}_t$ and the action $\vec{a}_t$. This can be expressed as
\begin{equation*}
 \vec{s}_{t+1} = (\vec{p}_{t+1}, \vec{w}_{t+1}, \vec{r}_{t+1}) = f_\text{fixed}( \vec{p}_t , \vec{w}_t,\vec{r}_t,\vec{z}_t, \vec{a}_t) = (\vec{p}_{t}, \vec{w}_{t}, f_\text{robot action}(\vec{r}_t, \vec{a}_t))
\end{equation*}

\item \textbf{Jacobian}: This baseline assumes that the object is rigidly attached to the environment. Therefore its pose w.r.t the environment does not change, i.e. $\vec{q}_{t+1} = \vec{q}_t$ and $\vec{w}_{t+1}=\vec{w}_t$. As \textit{Fixed Model}, future grasp frame poses $\vec{r}_{t+1}$ are predicted using a deterministic robot action model.
\begin{equation*}
 \vec{s}_{\text{obj}, t+1} = (\vec{q}_{t+1}, \vec{w}_{t+1}, \vec{r}_{t+1}) = f_\text{jacobian}( \vec{q}_t , \vec{w}_t,\vec{r}_t,\vec{z}_t, \vec{a}_t) = (\vec{q}_{t}, \vec{w}_{t}, f_\text{robot action}(\vec{r}_t, \vec{a}_t))
\end{equation*}
\item \textbf{Pseudo-random}: This baseline does not have a model at all. Instead, actions are randomly sampled from each task action space. Although the actions are random, they are constrained towards the goal. For drawing, pseudo-random actions constrain the robot motion to be toward the goal. For pivoting, the pseudo-random actions force motions that push against the table, i.e. actions that produce pivoting motions. This way we give an intuition about the task goal and prevent barren actions. We use rejection sampling to impose the constraints and discard actions that violate them.
\end{itemize}

\subsection{Membrane Dynamics vs Object Pose Dynamics}
\label{supp:membrane_vs_object_pose}

\begin{figure}[H]
    \centering
    \includegraphics[width=\textwidth]{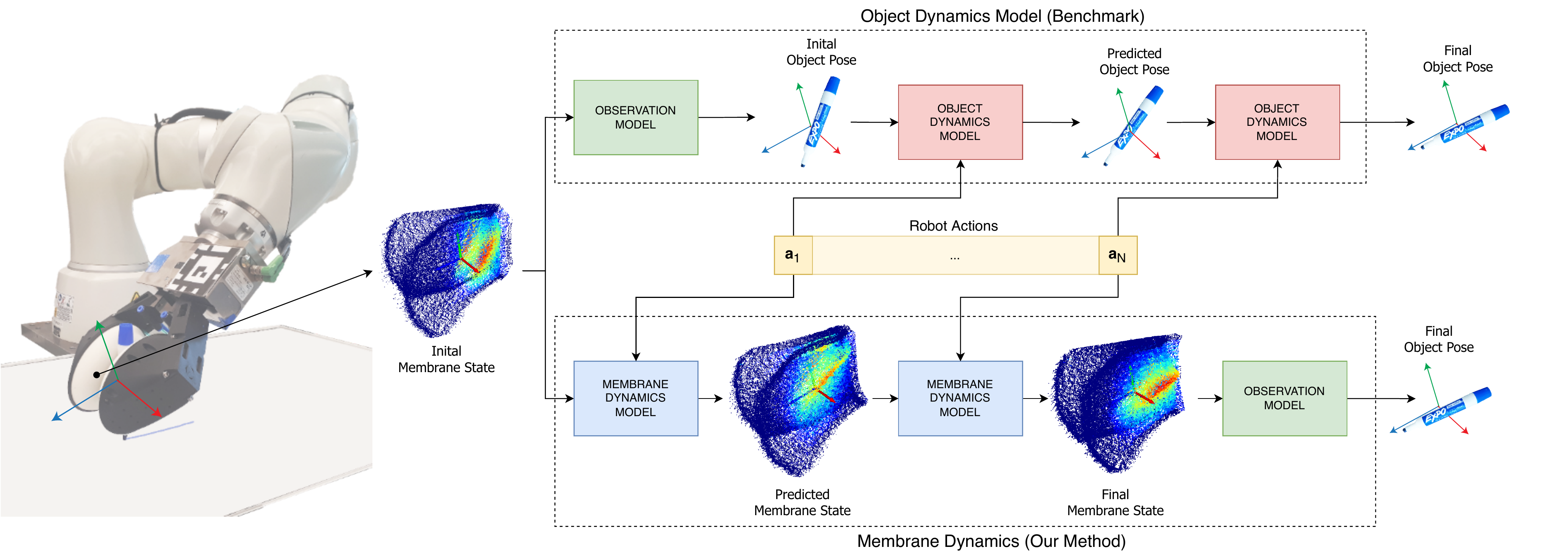}
    \caption{\textbf{Dynamics Model Comparison}  As opposed to extracting the current object pose and formulating dynamics in the object pose space (upper), our method (lower) characterizes the task dynamics as experienced by the tactile sensor by explicitly modeling the sensor's membrane deformations. As a result, our method's predicted dynamics are agnostic to the observation model.} 
    \label{fig:model_comparison}
\end{figure}

\subsection{Membrane Signature Processing}
\label{supp:membrane_singature_processing}

\begin{figure}[h!]
    \centering
    \includegraphics[width=\textwidth]{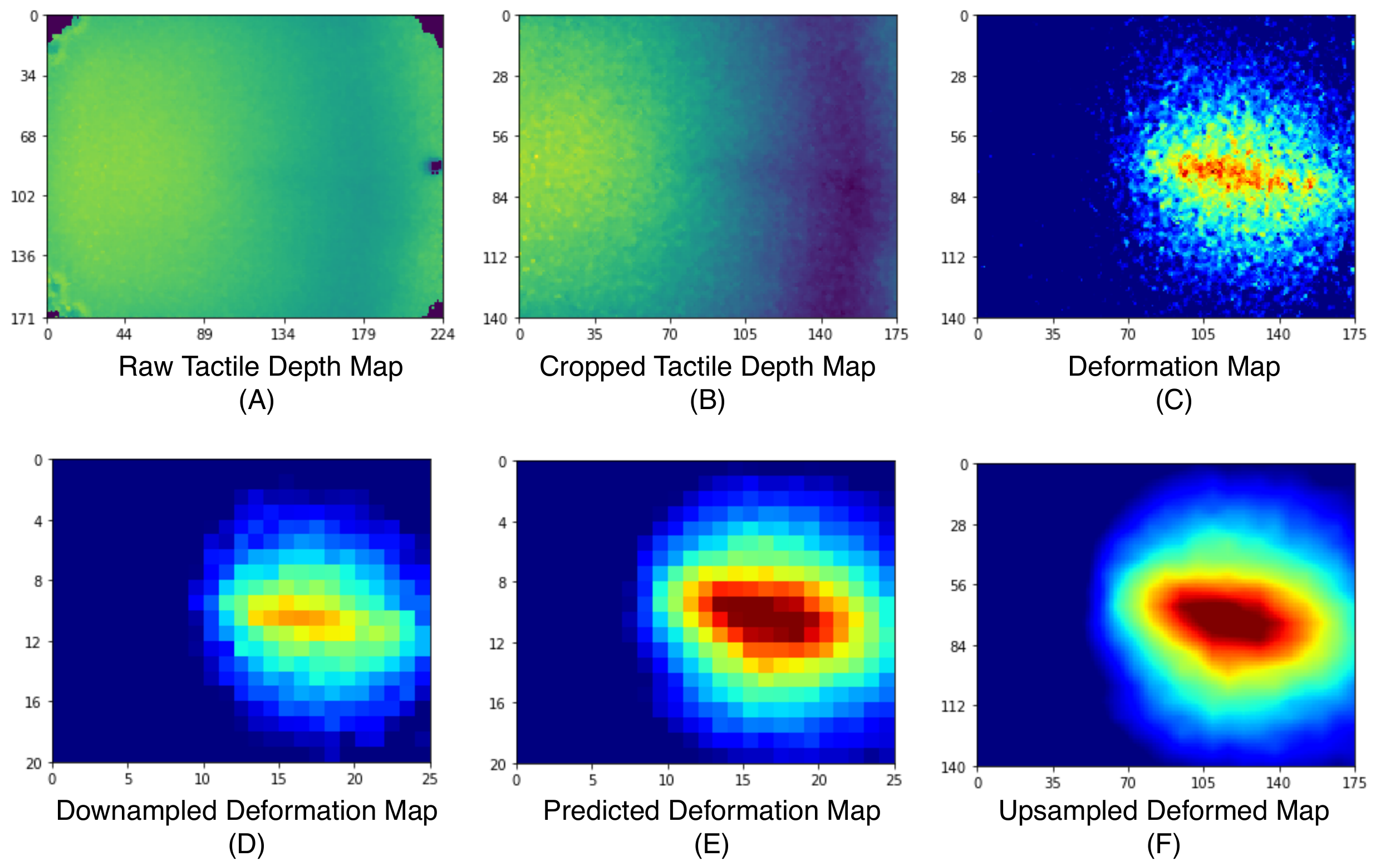}
    \caption{\textbf{Tactile Depth Map Signatures} The PicoFlexx camera's raw measurement (A) is processed (B) and compared with a reference state to extract the depth deformations (C). Our dynamics models operate on a downsampled the depth map (D), obtained using average pooling. Before applying the observation model on model predictions (E), we upsample the deformation map using bi-linear interpolation (F).}
    \label{fig:tactile_signatures}
\end{figure}

The raw tactile signature measured by the PMD PicoFlexx cameras is a $(224 \times 171)$ depth map (Fig. \ref{fig:tactile_signatures} A). We crop it to remove the borders and the noisy areas, obtaining a $(175 \times 140)$ depth map (Fig. \ref{fig:tactile_signatures} B).
Comparing the current measured signature with a reference signature of the undeformed sensors, we can obtain the deformation map $\vec{p}_t$ (Fig. \ref{fig:tactile_signatures} C), which expresses the relative change in camera distance for each pixel.
To simplify the model and the data requirements, our model works with a downsampled depth map. Using average-pooling with factors $(7,7)$, we reduce the deformation map size to $(25\times 20)$ (Fig. \ref{fig:tactile_signatures} D). This is the tactile signature size that the membrane dynamics model \ref{sec:membrane_dyn} takes as input.

\begin{equation*}
    \vec{p}_{t,\{L,R\}} = \vec{p}^\text{meas}_{t,\{L,R\}} - \vec{p}^\text{ref}_{t,\{L,R\}} \quad \text{where}\quad \vec{p}_{t,\{L,R\}} , \vec{p}^\text{meas}_{t,\{L,R\}} ,\vec{p}^\text{ref}_{t,\{L,R\}}  \in \R^{175\times 140}
\end{equation*}

The observation model expects tactile maps in the original resolution of $(175 \times 140)$. Therefore, predicted tactile state in the downsampled domain $\R^{25\times 20}$ (Fig. \ref{fig:tactile_signatures} E) need to be converted back to the original domain $\R^{175\times 140}$. To this end, we use bi-linear interpolation over the predicted deformation maps (Fig. \ref{fig:tactile_signatures} F).

\section{Experimental Setup}
\label{supp:experiment_setup}

For pivoting, the action space is composed by:
\begin{equation*}
\begin{split}
    \vec{a} = (\text{gw}, \Delta y, & \Delta z, \Delta\phi) \in \mathcal A_\text{pivoting}\\
    \text{where}\ \text{gw} &\in [5,40]\ \text{mm}\\
    \Delta y &\in [-40,40]\ \text{mm}\\
    \Delta z &\in [-d_\text{env}, 20]\ \text{mm}\\
    \Delta \phi &\in [-\frac{\pi}{6}, \frac{\pi}{6}]\  \text{rad}\\
\end{split}
\end{equation*}

For drawing, the action space is composed by:
\begin{equation*}
    \begin{split}
        \vec{a} = (\text{gw}, \Delta y, & \Delta z, \Delta\phi) \in \mathcal A_\text{drawing}\\
        \text{where}\ \text{gw} &\in [10,40]\ \text{mm}\\
    \Delta y &\in [0, 20]\ \text{mm}\\
    \Delta z &\in [-5, \alpha_\text{impedance}]\ \text{mm}\\
    \Delta \phi &\in [-\frac{\pi}{36}, \frac{\pi}{36}]\  \text{rad}\\
   \end{split} 
\end{equation*}

\subsection{Data Collection}
\label{supp:data_collection}
For each task we collect 4000 state-action-state transitions. Each transition sample is also conditioned on the tool geometry $\vec{z}_i$ of the grasped tool during the transition. As a consequence, each data sample is a quadruplet $(\state_t, \vec{z}_t, \vec{a}_t, \state_{t+1})$ . The data collection process depends on the task: 

\begin{itemize}
    \item \textbf{Drawing Data Collection}: For drawing, we collect samples combining random actions with epsilon-greedy samples from a Jacobian controller. We collect 2/3 of the data drawing the evaluation line. The robot starts at a fixed location and a fixed orientation. With probability $p=0.15$, the controller executes a random action. With probability $1-p$ the controller executes an action given by a Jacobian controller. The Jacobian controller assumes that the marker is rigidly attached to the robot. It outputs the action within the action space that will result in the object closest to the goal configuration of the marker vertically touching the board. This controller serves us as a baseline to bias the dataset towards states that are more likely to be found with the model and controller running.
    The remaining 1/3 of the data is collected by drawing lines purely random with different directions and starting points. We combine both data sources for a more rich dataset.
    \item \textbf{Pivoting Data Collection}: For pivoting, the data collection pipeline works as follows: First, the user feeds the tool to the robot and this brings it into contact with the environment with a relative orientaiton of $\pm45^\circ$. The sign is chosen randomly. Then, for a sequence of 5 steps, the robot performs random actions within the action space and records their initial and final states. If the tool falls out of hand, or the execution fails, the data point is discarded and the process is restarted.
\end{itemize}

\subsection{Drawing Evaluation Setup}
\label{supp:drawing_evaluation}
\begin{figure}[h!]
    \centering 
    \includegraphics[width=\textwidth]{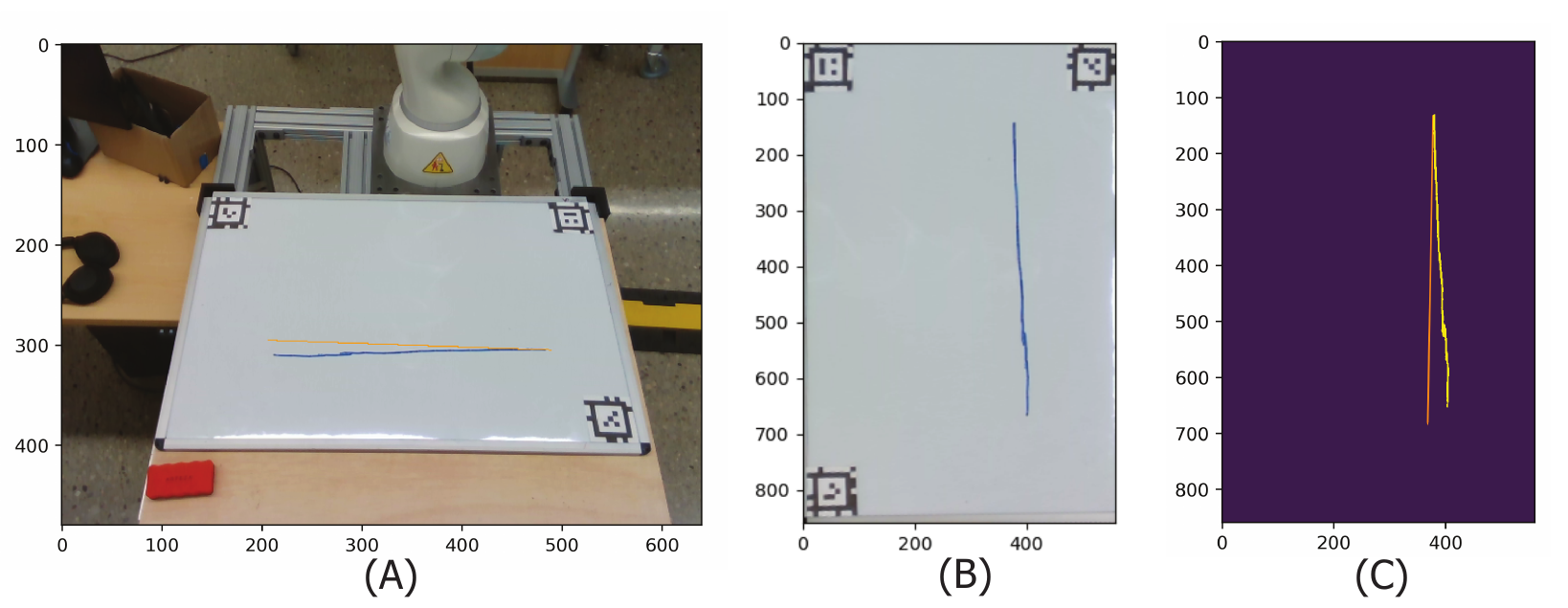}
    \caption{\textbf{Drawing Task Evaluation}
    (A) Actual drawing (blue) overlayed with the desired one (orange). (B) Unwarped board. (C) Unwarped detected drawing (yellow) compared with the desired drawing (orange). }
    \label{fig:drawing_evaluation}
\end{figure}

To evaluate the drawing accuracy, we mount a camera on a top-down view over the drawing board. The board has 3 AprilTag fiducial marks on its corners that track its position (Fig. \ref{fig:drawing_evaluation} A). The tags allow us to unwarp the board (Fig. \ref{fig:drawing_evaluation} B). Processing the unwarpped image and adaptive thresholding we can detect the actual drawing $\mat{D}^\text{meas}$ (Fig. \ref{fig:drawing_evaluation} C) and compare it with the desired one $\mat{D}^\text{goal}$.
Note that the resolution of the evaluation image is 1 pixel per millimeter.
The drawing score $\zeta_\text{drawing}$ is defined as follows:
\begin{equation*}
    \zeta_\text{drawing}(\mat{D}^\text{meas}, \mat{D}^\text{goal}) = \frac{\mat{D}^\text{meas} \cap \mat{D}^\text{goal}}{\mat{D}^\text{meas}}\in [0,1] \quad \text{where}\  \mat{D}^\text{meas},\mat{D}^\text{goal} \in \mathbb Z_2^{565 \times 860} 
\end{equation*}

\subsection{Pivoting Evaluation Setup}
To asses our algorithm on the pivoting task, we evaluate the accuracy of the achieved tool pose with respect to the desired pose. For simplicity, we only compare the tool orientation defined as the angle between the tool major axis and the $z$ axis of the grasp frame. 
The pivoting evaluation pipeline is initialized as follows: First, the user feeds the tool to the robot in an arbitrary angle within the range $[-45^\circ, 45^\circ]$. Second, a goal orientation is sampled from a uniform distribution in the range $[-150^\circ, 150^\circ]$. Third, the robot moves so the tool is oriented forming $\pm 45^\circ$ w.r.t. the environment surface. The sign is chosen depending on the direction of the necessary rotation towards the goal angle. L
Once the tool is in contact with the environment, the algorithm can execute up to 10 actions to drive the tool to the desired orientation. If the tool has reached the orientation within $\pm4^\circ$ of the goal orientation, we consider the execution successful. On the other side, if the tool slips out of grasp or it overshoots the goal angle, we consider a failure and the execution is stopped. We record the last final tool orientation. In case the tool has slipped out of hand, we report the last achieved stable pose.
For each tool we run 10 executions.
The pivoting score is computed as:
\begin{equation*}
    \zeta_\text{pivoting}(\theta^\text{achieved}, \theta^\text{goal}) = |\theta^\text{goal}- \theta^\text{achieved}| \in [0, 180^\circ]
\end{equation*}


\section{Supplementary Results}
\label{supp:supp_results}
\subsection{Pivoting Supplementary Results}
\label{supp:pivoting_results}
We evaluate the pivoting performance based on the difference angle between the achieved tool angle and the desired tool angle. The pivoting score is therefore the difference between these two values, which ideally should be 0. We use our observation model to estimate the final reached pose. 

The desired tool pose is sampled at random within the graspable range.  The controller optimizes the pivoting action sequence to achieve that final pose exploiting the dynamics model.

\begin{figure}[h!]
    \includegraphics[width=\linewidth]{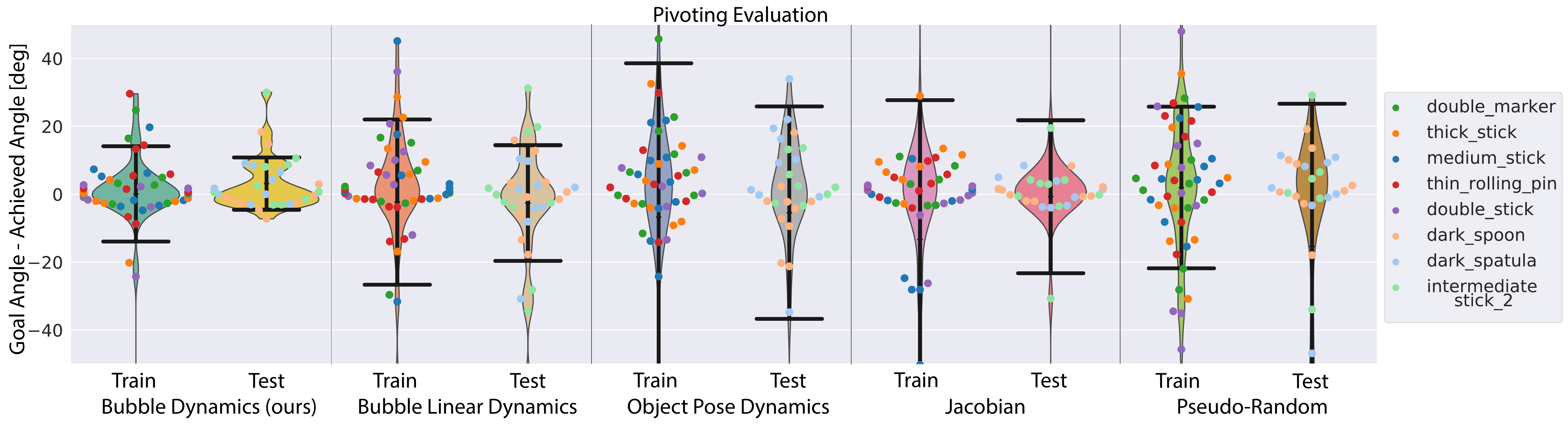}
    \vspace{-15pt}
    \label{fig:pivoting_evaluation}
    \caption{\textbf{Pivoting Evaluation Results} Each column pair represents an evaluated method on both train tools and unseen tools. Colored dots indicate each tool achieved score. Black horizontal lines indicate the sample std around the mean. The pivoting score is the difference in degrees between the desired tool angle and the achieved tool angle. Therefore, the optimal pivoting score is 0. }

\end{figure}



\end{appendices}

%% file: main.bbl
\begin{thebibliography}{36}
\providecommand{\natexlab}[1]{#1}
\providecommand{\url}[1]{\texttt{#1}}
\expandafter\ifx\csname urlstyle\endcsname\relax
  \providecommand{\doi}[1]{doi: #1}\else
  \providecommand{\doi}{doi: \begingroup \urlstyle{rm}\Url}\fi

\bibitem[Rodriguez(2021)]{rodriguez2021unstable}
A.~Rodriguez.
\newblock The unstable queen: Uncertainty, mechanics, and tactile feedback.
\newblock \emph{Science Robotics}, 6\penalty0 (54):\penalty0 eabi4667, 2021.

\bibitem[Alspach et~al.(2019)Alspach, Hashimoto, Kuppuswamy, and
  Tedrake]{alspach2019soft}
A.~Alspach, K.~Hashimoto, N.~Kuppuswamy, and R.~Tedrake.
\newblock Soft-bubble: A highly compliant dense geometry tactile sensor for
  robot manipulation.
\newblock In \emph{2019 2nd IEEE International Conference on Soft Robotics
  (RoboSoft)}, pages 597--604. IEEE, 2019.

\bibitem[Donlon et~al.(2018)Donlon, Dong, Liu, Li, Adelson, and
  Rodriguez]{donlon2018gelslim}
E.~Donlon, S.~Dong, M.~Liu, J.~Li, E.~Adelson, and A.~Rodriguez.
\newblock Gelslim: A high-resolution, compact, robust, and calibrated
  tactile-sensing finger.
\newblock In \emph{2018 IEEE/RSJ International Conference on Intelligent Robots
  and Systems (IROS)}, pages 1927--1934. IEEE, 2018.

\bibitem[Lambeta et~al.(2020)Lambeta, Chou, Tian, Yang, Maloon, Most, Stroud,
  Santos, Byagowi, Kammerer, et~al.]{lambeta2020digit}
M.~Lambeta, P.-W. Chou, S.~Tian, B.~Yang, B.~Maloon, V.~R. Most, D.~Stroud,
  R.~Santos, A.~Byagowi, G.~Kammerer, et~al.
\newblock Digit: A novel design for a low-cost compact high-resolution tactile
  sensor with application to in-hand manipulation.
\newblock \emph{IEEE Robotics and Automation Letters}, 5\penalty0 (3):\penalty0
  3838--3845, 2020.

\bibitem[Yamaguchi(2018)]{yamaguchi2018fingervision}
A.~Yamaguchi.
\newblock Fingervision for tactile behaviors, manipulation, and haptic feedback
  teleoperation.
\newblock In \emph{the 4th IEEJ international workshop on Sensing, Actuation,
  Motion Control, and Optimization (SAMCON2018)}, 2018.

\bibitem[Bauza et~al.(2022)Bauza, Bronars, and Rodriguez]{bauza2022tac2pose}
M.~Bauza, A.~Bronars, and A.~Rodriguez.
\newblock Tac2pose: Tactile object pose estimation from the first touch.
\newblock \emph{arXiv preprint arXiv:2204.11701}, 2022.

\bibitem[Bauza et~al.(2020)Bauza, Valls, Lim, Sechopoulos, and
  Rodriguez]{bauzaFirstTouch}
M.~Bauza, E.~Valls, B.~Lim, T.~Sechopoulos, and A.~Rodriguez.
\newblock Tactile object pose estimation from the first touch with geometric
  contact rendering.
\newblock \emph{arXiv preprint arXiv:2012.05205}, 2020.

\bibitem[Ma et~al.(2021)Ma, Dong, and Rodriguez]{daolinGelslimContactSensing}
D.~Ma, S.~Dong, and A.~Rodriguez.
\newblock Extrinsic contact sensing with relative-motion tracking from
  distributed tactile measurements.
\newblock In \emph{2021 IEEE International Conference on Robotics and
  Automation (ICRA)}, pages 11262--11268, 2021.
\newblock \doi{10.1109/ICRA48506.2021.9561781}.

\bibitem[Kuppuswamy et~al.(2020)Kuppuswamy, Castro, Phillips-Grafflin, Alspach,
  and Tedrake]{bubblesStateEstimation}
N.~Kuppuswamy, A.~Castro, C.~Phillips-Grafflin, A.~Alspach, and R.~Tedrake.
\newblock Fast model-based contact patch and pose estimation for highly
  deformable dense-geometry tactile sensors.
\newblock \emph{IEEE Robotics and Automation Letters}, 5\penalty0 (2):\penalty0
  1811--1818, 2020.
\newblock \doi{10.1109/LRA.2019.2961050}.

\bibitem[Kelestemur et~al.(2022)Kelestemur, Platt, and
  Padir]{kelestemur2022tactile}
T.~Kelestemur, R.~Platt, and T.~Padir.
\newblock Tactile pose estimation and policy learning for unknown object
  manipulation.
\newblock \emph{arXiv preprint arXiv:2203.10685}, 2022.

\bibitem[Narang et~al.(2021)Narang, Sundaralingam, Van~Wyk, Mousavian, and
  Fox]{narang2021interpreting}
Y.~S. Narang, B.~Sundaralingam, K.~Van~Wyk, A.~Mousavian, and D.~Fox.
\newblock Interpreting and predicting tactile signals for the syntouch biotac.
\newblock \emph{arXiv preprint arXiv:2101.05452}, 2021.

\bibitem[Li et~al.(2013)Li, Sch{\"u}rmann, Haschke, and Ritter]{li2013control}
Q.~Li, C.~Sch{\"u}rmann, R.~Haschke, and H.~J. Ritter.
\newblock A control framework for tactile servoing.
\newblock In \emph{Robotics: Science and systems}. Citeseer, 2013.

\bibitem[Lepora et~al.(2017)Lepora, Aquilina, and
  Cramphorn]{lepora2017exploratory}
N.~F. Lepora, K.~Aquilina, and L.~Cramphorn.
\newblock Exploratory tactile servoing with active touch.
\newblock \emph{IEEE Robotics and Automation Letters}, 2\penalty0 (2):\penalty0
  1156--1163, 2017.

\bibitem[Tian et~al.(2019)Tian, Ebert, Jayaraman, Mudigonda, Finn, Calandra,
  and Levine]{tactileMPC}
S.~Tian, F.~Ebert, D.~Jayaraman, M.~Mudigonda, C.~Finn, R.~Calandra, and
  S.~Levine.
\newblock Manipulation by feel: Touch-based control with deep predictive
  models.
\newblock In \emph{2019 International Conference on Robotics and Automation
  (ICRA)}, pages 818--824. IEEE, 2019.

\bibitem[Van~Hoof et~al.(2016)Van~Hoof, Chen, Karl, van~der Smagt, and
  Peters]{van2016stable}
H.~Van~Hoof, N.~Chen, M.~Karl, P.~van~der Smagt, and J.~Peters.
\newblock Stable reinforcement learning with autoencoders for tactile and
  visual data.
\newblock In \emph{2016 IEEE/RSJ international conference on intelligent robots
  and systems (IROS)}, pages 3928--3934. IEEE, 2016.

\bibitem[De~Luca et~al.(2006)De~Luca, Albu-Schaffer, Haddadin, and
  Hirzinger]{de2006collision}
A.~De~Luca, A.~Albu-Schaffer, S.~Haddadin, and G.~Hirzinger.
\newblock Collision detection and safe reaction with the dlr-iii lightweight
  manipulator arm.
\newblock In \emph{2006 IEEE/RSJ International Conference on Intelligent Robots
  and Systems}, pages 1623--1630. IEEE, 2006.

\bibitem[Magrini et~al.(2014)Magrini, Flacco, and
  De~Luca]{magrini2014estimation}
E.~Magrini, F.~Flacco, and A.~De~Luca.
\newblock Estimation of contact forces using a virtual force sensor.
\newblock In \emph{2014 IEEE/RSJ International Conference on Intelligent Robots
  and Systems}, pages 2126--2133, 2014.
\newblock URL \url{IEEE}.

\bibitem[Manuelli and Tedrake(2016)]{manuelli2016localizing}
L.~Manuelli and R.~Tedrake.
\newblock Localizing external contact using proprioceptive sensors: The contact
  particle filter.
\newblock In \emph{2016 IEEE/RSJ International Conference on Intelligent Robots
  and Systems (IROS)}, pages 5062--5069. IEEE, 2016.

\bibitem[bio(2018)]{biotak}
Biotac sensor, syntouch inc. sensor technology.
\newblock In \emph{[Online; accessed 2022 Jan 15]}. Available from:
  https://www.syntouchinc. com/sensor-technology/, 2018.

\bibitem[Yamaguchi and Atkeson(2019)]{yamaguchi2019recent}
A.~Yamaguchi and C.~G. Atkeson.
\newblock Recent progress in tactile sensing and sensors for robotic
  manipulation: can we turn tactile sensing into vision?
\newblock \emph{Advanced Robotics}, 33\penalty0 (14):\penalty0 661--673, 2019.

\bibitem[Wang et~al.(2022)Wang, Lambeta, Chou, and Calandra]{tacto}
S.~Wang, M.~Lambeta, P.-W. Chou, and R.~Calandra.
\newblock {TACTO}: A fast, flexible, and open-source simulator for
  high-resolution vision-based tactile sensors.
\newblock \emph{{IEEE} Robotics and Automation Letters}, 7\penalty0
  (2):\penalty0 3930--3937, apr 2022.
\newblock \doi{10.1109/lra.2022.3146945}.
\newblock URL \url{https://doi.org/10.1109%2Flra.2022.3146945}.

\bibitem[Kelestemur et~al.(2022)Kelestemur, Platt, and Padir]{tarik2022}
T.~Kelestemur, R.~Platt, and T.~Padir.
\newblock Tactile pose estimation and policy learning for unknown object
  manipulation, 2022.
\newblock URL \url{https://arxiv.org/abs/2203.10685}.

\bibitem[Hogan et~al.(2020)Hogan, Ballester, Dong, and
  Rodriguez]{hogan2020tactile}
F.~R. Hogan, J.~Ballester, S.~Dong, and A.~Rodriguez.
\newblock Tactile dexterity: Manipulation primitives with tactile feedback.
\newblock In \emph{2020 IEEE international conference on robotics and
  automation (ICRA)}, pages 8863--8869. IEEE, 2020.

\bibitem[She et~al.(2021)She, Wang, Dong, Sunil, Rodriguez, and
  Adelson]{she2021cable}
Y.~She, S.~Wang, S.~Dong, N.~Sunil, A.~Rodriguez, and E.~Adelson.
\newblock Cable manipulation with a tactile-reactive gripper.
\newblock \emph{The International Journal of Robotics Research}, 40\penalty0
  (12-14):\penalty0 1385--1401, 2021.

\bibitem[Kim and Rodriguez(2021)]{kim2021active}
S.~Kim and A.~Rodriguez.
\newblock Active extrinsic contact sensing: Application to general peg-in-hole
  insertion.
\newblock \emph{arXiv preprint arXiv:2110.03555}, 2021.

\bibitem[Qi et~al.(2017)Qi, Su, Mo, and Guibas]{pointnet}
C.~R. Qi, H.~Su, K.~Mo, and L.~J. Guibas.
\newblock Pointnet: Deep learning on point sets for 3d classification and
  segmentation.
\newblock In \emph{Proceedings of the IEEE conference on computer vision and
  pattern recognition}, pages 652--660, 2017.

\bibitem[Wu et~al.(2015)Wu, Song, Khosla, Yu, Zhang, Tang, and
  Xiao]{modelnet40}
Z.~Wu, S.~Song, A.~Khosla, F.~Yu, L.~Zhang, X.~Tang, and J.~Xiao.
\newblock 3d shapenets: A deep representation for volumetric shapes.
\newblock In \emph{Proceedings of the IEEE conference on computer vision and
  pattern recognition}, pages 1912--1920, 2015.

\bibitem[Rumelhart et~al.(1985)Rumelhart, Hinton, and
  Williams]{rumelhart1985learning}
D.~E. Rumelhart, G.~E. Hinton, and R.~J. Williams.
\newblock Learning internal representations by error propagation.
\newblock Technical report, California Univ San Diego La Jolla Inst for
  Cognitive Science, 1985.

\bibitem[Koval et~al.(2015)Koval, Pollard, and Srinivasa]{koval2015pose}
M.~C. Koval, N.~S. Pollard, and S.~S. Srinivasa.
\newblock Pose estimation for planar contact manipulation with manifold
  particle filters.
\newblock \emph{The International Journal of Robotics Research}, 34\penalty0
  (7):\penalty0 922--945, 2015.

\bibitem[Williams et~al.(2015)Williams, Aldrich, and Theodorou]{mppi1}
G.~Williams, A.~Aldrich, and E.~Theodorou.
\newblock Model predictive path integral control using covariance variable
  importance sampling.
\newblock \emph{arXiv preprint arXiv:1509.01149}, 2015.

\bibitem[Hirose et~al.(2019)Hirose, Xia, Mart{\'\i}n-Mart{\'\i}n, Sadeghian,
  and Savarese]{hirose2019deep}
N.~Hirose, F.~Xia, R.~Mart{\'\i}n-Mart{\'\i}n, A.~Sadeghian, and S.~Savarese.
\newblock Deep visual mpc-policy learning for navigation.
\newblock \emph{IEEE Robotics and Automation Letters}, 4\penalty0 (4):\penalty0
  3184--3191, 2019.

\bibitem[Power and Berenson(2022)]{power2022variational}
T.~Power and D.~Berenson.
\newblock Variational inference mpc using normalizing flows and
  out-of-distribution projection.
\newblock \emph{arXiv preprint arXiv:2205.04667}, 2022.

\bibitem[Pfrommer et~al.(2020)Pfrommer, Halm, and
  Posa]{pfrommer2020contactnets}
S.~Pfrommer, M.~Halm, and M.~Posa.
\newblock Contactnets: Learning discontinuous contact dynamics with smooth,
  implicit representations.
\newblock \emph{arXiv preprint arXiv:2009.11193}, 2020.

\bibitem[Taylor et~al.(2021)Taylor, Dong, and Rodriguez]{taylor2021gelslim3}
I.~Taylor, S.~Dong, and A.~Rodriguez.
\newblock Gelslim3. 0: High-resolution measurement of shape, force and slip in
  a compact tactile-sensing finger.
\newblock \emph{arXiv preprint arXiv:2103.12269}, 2021.

\bibitem[Farneb{\"a}ck(2003)]{farneback2003two}
G.~Farneb{\"a}ck.
\newblock Two-frame motion estimation based on polynomial expansion.
\newblock In \emph{Scandinavian conference on Image analysis}, pages 363--370.
  Springer, 2003.

\bibitem[UM-ARM-Lab()]{pytorch_mppi}
UM-ARM-Lab.
\newblock Pytorch mppi.
\newblock URL \url{https://github.com/UM-ARM-Lab/pytorch_mppi}.

\end{thebibliography}
